\documentclass[journal,romanappendices]{IEEEtran}
\ifCLASSINFOpdf
  \usepackage[pdftex]{graphicx}
\else
\fi
\usepackage{amsmath}
\usepackage{algorithmic}

\usepackage{array}

\ifCLASSOPTIONcompsoc
 \usepackage[caption=false,font=normalsize,labelfont=sf,textfont=sf]{subfig}
\else
 \usepackage[caption=false,font=footnotesize]{subfig}
\fi
\usepackage{url}

\usepackage{color}
\newcolumntype{M}[1]{>{\centering\arraybackslash}m{#1}}
\usepackage{capt-of}
\usepackage{float}
\usepackage{enumitem}
\usepackage{multirow}
\newsavebox{\ieeealgbox}
\newenvironment{boxedalgorithmic}
  {\begin{lrbox}{\ieeealgbox}
   \begin{minipage}{\dimexpr 2\columnwidth-2\fboxsep-2\fboxrule}
   \begin{algorithmic}}
  {\end{algorithmic}
   \end{minipage}
 \end{lrbox}\noindent\fbox{\usebox{\ieeealgbox}}}

\begin{document}
%
\title{Semantic Labeling of Large-Area Geographic Regions
  Using Multi-View and Multi-Date Satellite Images and
  Noisy OSM Training Labels}
%
%
%

\author{Bharath~Comandur and Avinash C.
  Kak
  \thanks{This work was supported by the Intelligence
    Advanced Research Projects Activity (IARPA) via
    Department of Interior / Interior Business Center
    (DOI/IBC) contract number D17PC00280. The
    U.S. Government is authorized to reproduce and
    distribute reprints for Governmental purposes
    notwithstanding any copyright annotation
    thereon. Disclaimer: The views and conclusions contained
    herein are those of the authors and should not be
    interpreted as necessarily representing the official
    policies or endorsements, either expressed or implied,
    of IARPA, DOI/IBC, or the U.S. Government.}
  \thanks{The authors are with the Department of Electrical
    and Computer Engineering, Purdue University, West
    Lafayette, IN, 47907 USA (e-mail: bcomandu@purdue.edu;
    kak@purdue.edu).}

}

\maketitle

\begin{abstract}

  We present a novel multi-view training framework and CNN
  architecture for combining information from multiple
  overlapping satellite images and noisy training labels
  derived from OpenStreetMap (OSM) to semantically label
  buildings and roads across large geographic regions (100
  km$^2$). Our approach to multi-view semantic segmentation
  yields a 4-7\% improvement in the per-class IoU scores
  compared to the traditional approaches that use the views
  independently of one another.  A unique (and, perhaps,
  surprising) property of our system is that
  modifications that are added to the tail-end of the CNN
  for learning from the multi-view data can be discarded at
  the time of inference with a relatively small penalty in
  the overall performance. This implies that the benefits of
  training using multiple views are absorbed by all the
  layers of the network.
  Additionally, our approach only adds a small overhead in
  terms of the GPU-memory consumption even when training
  with as many as 32 views per scene.  The system we present
  is end-to-end automated, which facilitates comparing the
  classifiers trained directly on true orthophotos
  vis-a-vis first training them on the off-nadir images and
  subsequently translating the predicted labels to
  geographical coordinates.  \textit{With no human
    supervision}, our IoU scores for the buildings and roads
  classes are 0.8 and 0.64 respectively which are better
  than state-of-the-art approaches that use OSM labels and
  that are not completely automated.
  
\end{abstract}

\begin{IEEEkeywords}
  Multi-View Semantic Segmentation, OSM, Deep Learning, CNN,
  Noisy Labels, DSM
\end{IEEEkeywords}

%
\IEEEpeerreviewmaketitle

\section{Introduction}

 

\IEEEPARstart{I}{n} this work, we are interested in
answering the following question -- {\em Is there an optimal
  way to combine multi-view and multi-date satellite images,
  and noisy training labels derived from OpenStreetMap (OSM)
  \cite{osm} for the task of semantically labeling
  buildings and roads on the ground over large geographic
  regions (100 km$^2$)}? Note that labeling points on the
ground is more challenging than labeling pixels in images
because the former requires that we first map each point on
the ground to the correct pixel in each image. This is only
possible if (1) the multi-date and multi-view images are
not only aligned with one another but are also aligned well
in an absolute sense to the real world; and (2) if we have
accurate knowledge of the heights of the points on the
ground.

Before summarizing our main contributions, to give the
reader a glimpse of the power of the approach presented in
this study, we show some sample results in
Fig. \ref{fig:building-sv-mv}.

\begin{table*}[h]
  \setlength{\tabcolsep}{0.07cm}
  \begin{center}
      \begin{tabular}{|M{1.4cm}|M{0.22\linewidth}|M{0.22\linewidth}|M{0.22\linewidth}|M{0.22\linewidth}|}
      \hline
      Single-View Training (Baseline) & \raisebox{-.5\height}{\includegraphics[width=1\linewidth, height = 1\linewidth]{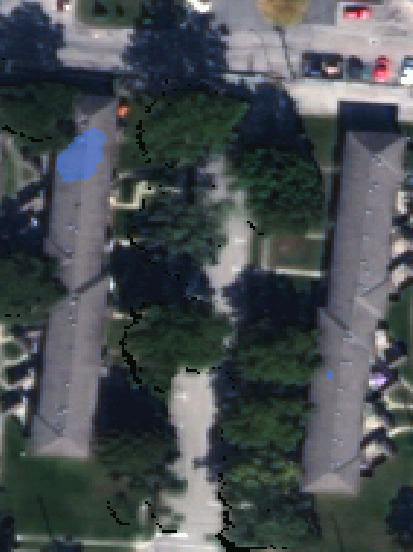}}
      & \raisebox{-.5\height}{\includegraphics[width=1\linewidth, height = 1\linewidth]{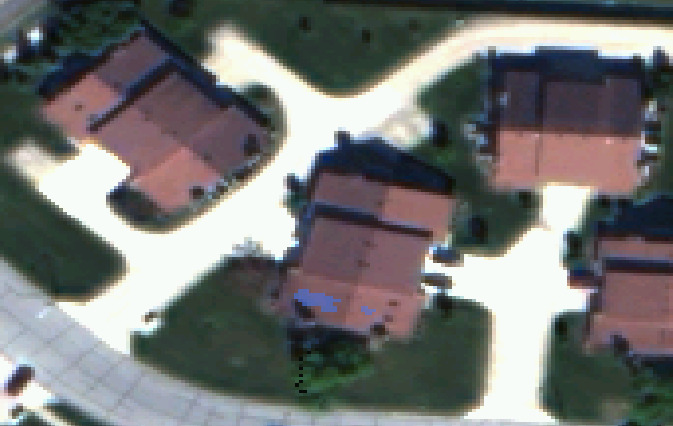}} 
      & \raisebox{-.5\height}{\includegraphics[width=1\linewidth, height = 1\linewidth]{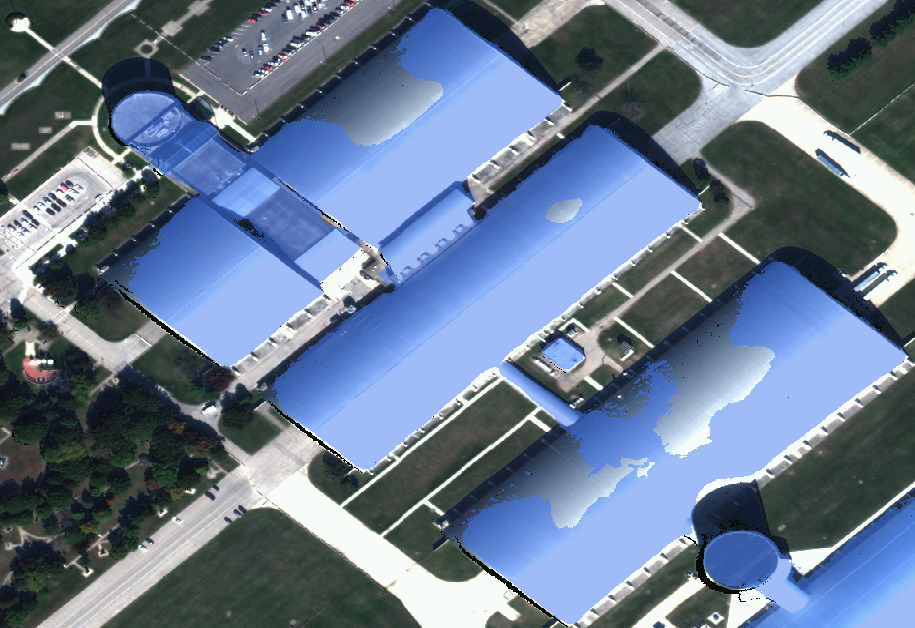}} 
      & \raisebox{-.5\height}{\includegraphics[width=1\linewidth, height = 1\linewidth]{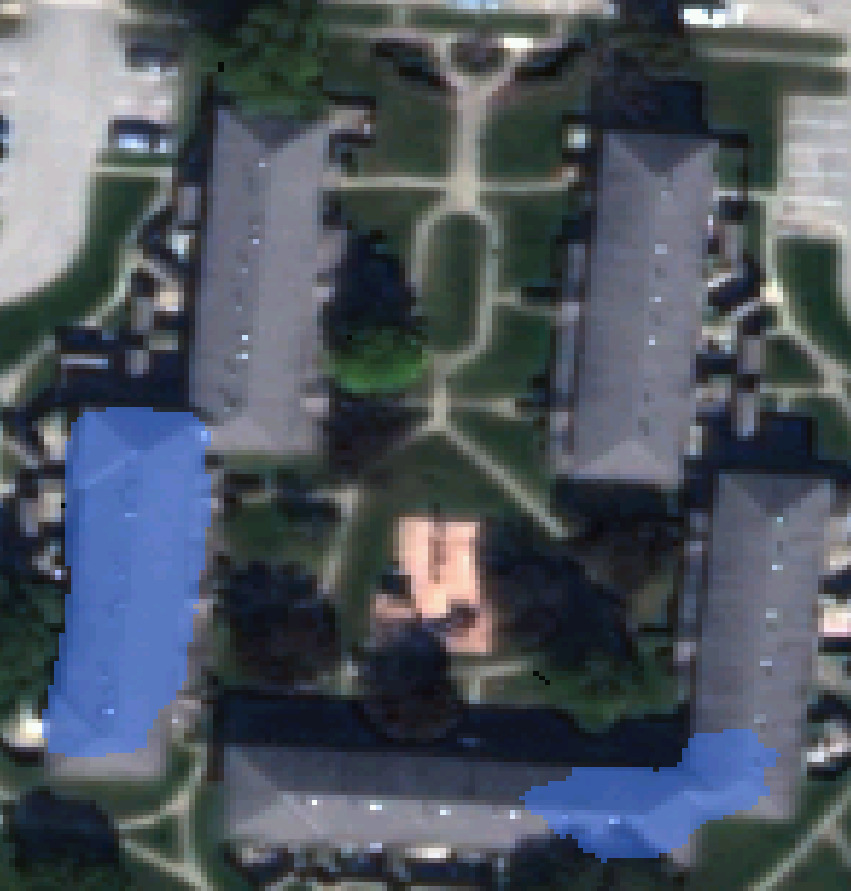}} \\
      \hline
      \hline
      Multi-View Training (Proposed Approach) & \raisebox{-.5\height}{\includegraphics[width=1\linewidth, height = 1\linewidth]{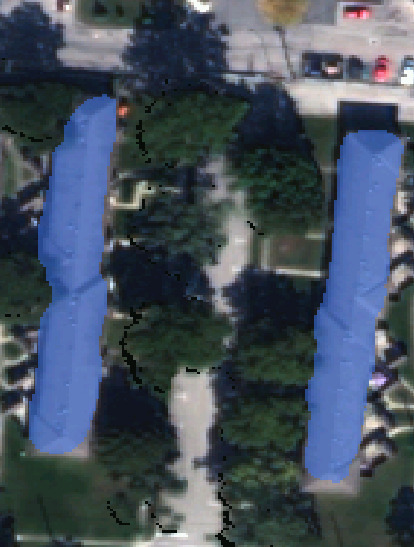}} 
      & \raisebox{-.5\height}{\includegraphics[width=1\linewidth, height = 1\linewidth]{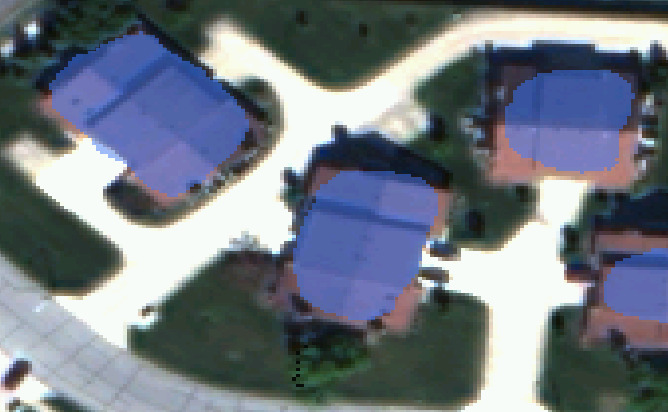}}
      & \raisebox{-.5\height}{\includegraphics[width=1\linewidth, height = 1\linewidth]{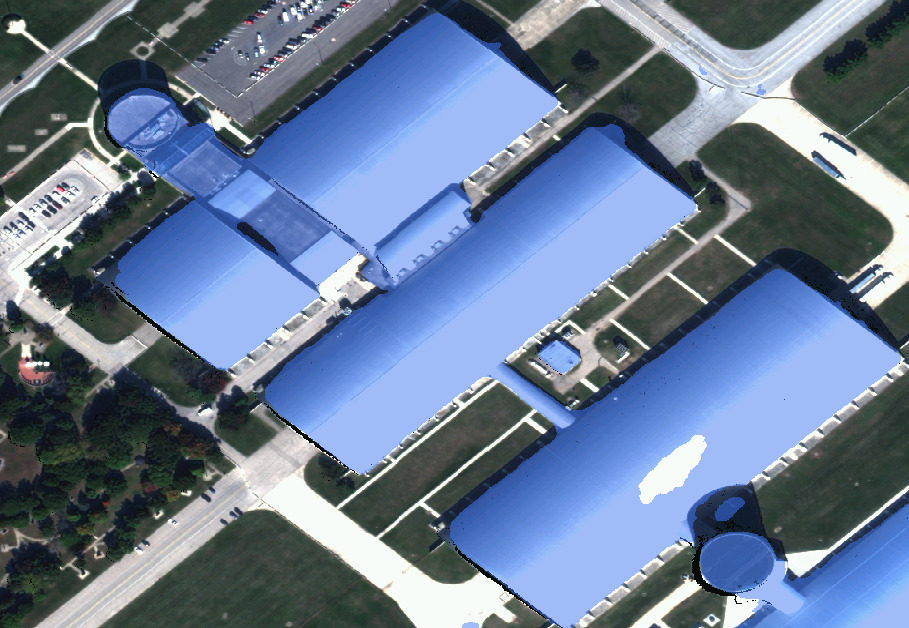}}
      & \raisebox{-.5\height}{\includegraphics[width=1\linewidth, height = 1\linewidth]{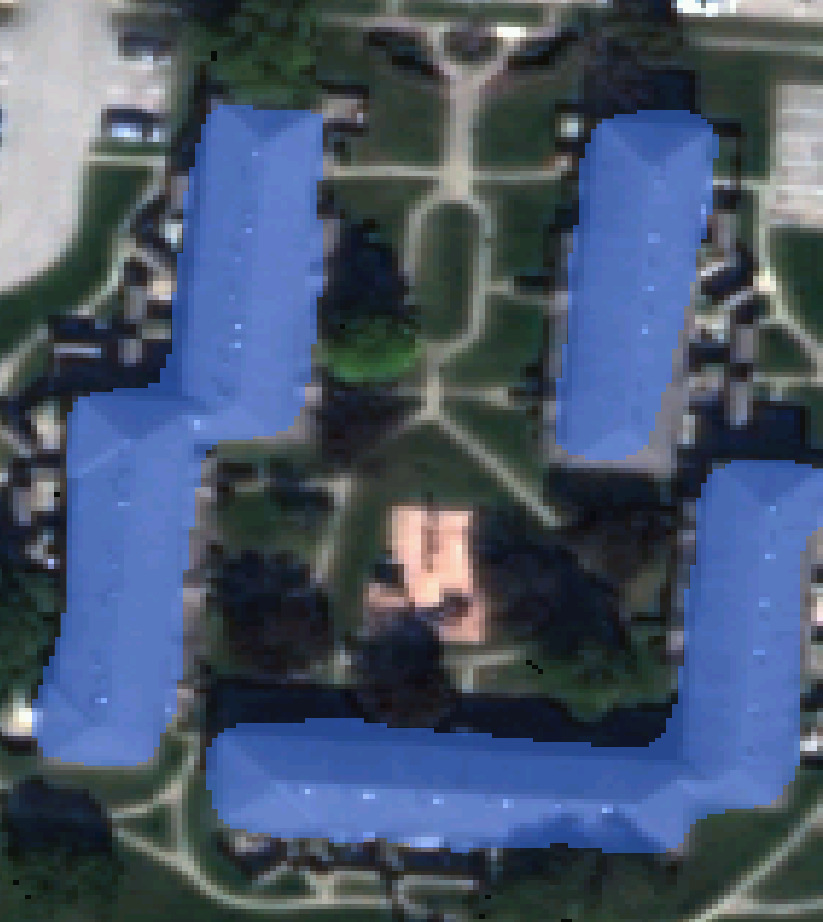}} \\
      \hline
    \end{tabular}
  \end{center}
  \captionof{figure}{To illustrate the power of our
    approach, the buildings in the bottom row were extracted
    by our approach based on multi-view training for
    semantic labeling. Compare with the top row where the
    training is based on single-views. Building points are
    marked in translucent \textcolor{blue}{blue}.}
  \label{fig:building-sv-mv}
\end{table*}

Towards answering the aforementioned question, we put forth
the following contributions:
\begin{enumerate}
\setlength\itemsep{1pt}
\item We present a novel multi-view training paradigm that
  yields improvements in the range \textit{4-7\% in the
    per-class IoU} (Intersection over Union)
  metric. \textit{ Our evaluation directly demonstrates that
    updating the weights of the convolutional neural network
    (CNN) by simultaneously learning from multiple views of
    the same scene can help alleviate the burden of noisy
    training labels.}

\item We present a direct comparison between training
  classifiers on 8-band true orthophoto images vis-a-vis
  training them on the original off-nadir images captured by
  the satellites. The fact that we use OSM training labels
  poses challenges for the latter approach, as it necessitates
  the need to transform labels from geographic coordinates
  into the off-nadir image-pixel coordinates. Such a
  transformation requires that we have knowledge of the
  heights of the points. The comparison presented in this
  study is unlike most published work in the literature that
  use pre-orthorectified single-view images.
  Additionally, we have \textit{released our software for creating
  true orthophotos, for public use}. Interested researchers
  can download this software from the link at \cite{gwarp}.

\item In order to make the above comparison possible, we
  present a true end-to-end automated framework that aligns
  large multi-view, multi-date images (each containing about
  $43008 \times 38000$ pixels), constructs a high-resolution
  accurate Digital Surface Model (DSM) over a 100 km$^2$
  area (which is needed for establishing correspondences
  between the pixels in the off-nadir images and points on
  the ground), and learns from noisy OSM labels \textbf{
    without any additional human supervision}.

\end{enumerate}

\begin{figure*}[h]
  \centering \subfloat[]
  {
    \includegraphics[width=0.47\linewidth,height=0.4\linewidth]{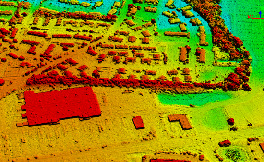}
    \label{fig:dsm_ohio_1}
  } \subfloat[]
  {
    \includegraphics[width=0.47\linewidth,height=0.4\linewidth]{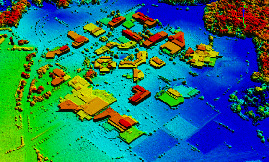}
    \label{fig:dsm_ohio_2}
  }\\
  \subfloat[]
  {
    \includegraphics[width=0.47\linewidth,height=0.4\linewidth]{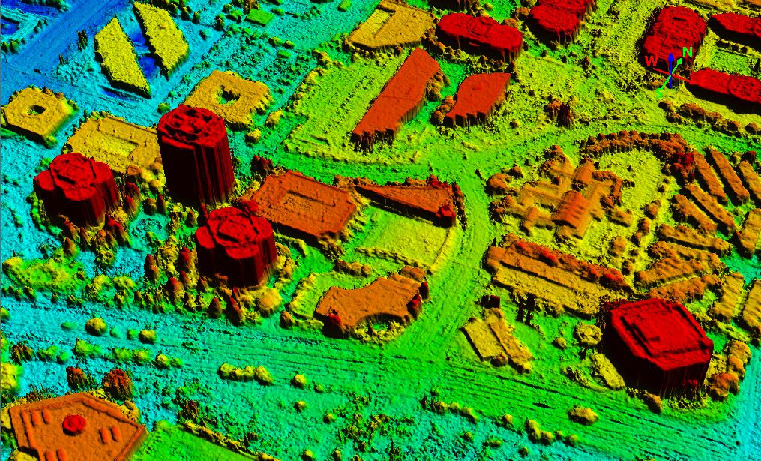}
    \label{fig:dsm_cali_1}
  }
  \subfloat[]
  {
    \includegraphics[width=0.47\linewidth,height=0.4\linewidth]{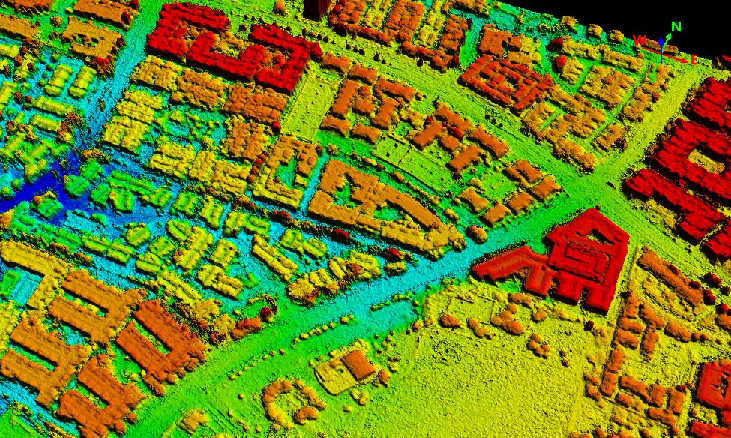}
    \label{fig:dsm_cali_2}
  }
  \caption{We have uploaded as Supporting Material the
    \textbf{flyby videos} and the images of the DSMs for two
    large areas, a 120 {\boldmath km$^2$} area from Ohio and
    a 62 {\boldmath km$^2$} area from California. The flyby
    videos can also be viewed at the link at
    \cite{flyby}. The top two images depict two small
    sections from the Ohio DSM, and the bottom two images depict two small sections
    from the California DSM. The DSM depictions have been
    colored according to the elevation values within the
    boundaries of each section.}

  \label{fig:dsm_example}
\end{figure*}

For our study, we use WorldView-3 (WV3) \cite{WV3} images
collected over two regions in Ohio and California, USA. We
use 32 images for each region. The images were collected
across a span of 2 years under varying conditions. Automatic
alignment and DSM construction are carried out for both
regions.  Smaller sections of these DSMs are shown in
Fig. \ref{fig:dsm_example}.

The rest of this manuscript is organized as follows. In Section
\ref{sec:lit_review}, we briefly review relevant
literature. Section \ref{sec:system_overview} provides
details on aligning images, creating large-area DSMs, and
deriving training labels from OSM. Section
\ref{sec:mvapproach} presents different approaches for
training and inference using CNNs. Section
\ref{sec:offnadir} discusses a strategy for using training
labels derived from OSM to label off-nadir
images. Experimental evaluation is described in Section
\ref{sec:evaluation}. Concluding remarks are presented in
Section \ref{sec:conclusion}.

\section{Literature Review}
\label{sec:lit_review} 

State-of-the-art approaches that demonstrate the use of
labels derived from OSM for finding roads and/or buildings
in overhead images include the studies described in
\cite{polymapper}, \cite{roadtracer},
\cite{bastani2018machine}, \cite{Chu_2019_ICCV},
\cite{yang2019road}, \cite{park2019refining},
\cite{Etten_2020_WACV}, \cite{mosinska2019joint},
\cite{yang2019road_v2}, \cite{eth_cnn}, \cite{resunet},
\cite{saito}, \cite{forez}, \cite{mnih_hybrid},
\cite{DeepVGI}, \cite{OSMDeepOD} and
\cite{DeepOSM}. Many of these approaches use some
  category of neural networks as part of their
  machine-learning frameworks. For instance, while the study
  described in \cite{polymapper} uses a CNN backbone to
  extract keypoints that are subsequently input to a
  recurrent neural network (RNN) to extract building
  polygons and road networks, the approach presented in
  \cite{roadtracer} constructs a road network in an
  iterative fashion by using a CNN to detect the next road
  segment given the previously extracted road network. The
  work discussed in \cite{park2019refining} builds upon the
  approach in \cite{roadtracer} by using a generative
  adversarial network (GAN) \cite{gans} to further refine
  the outputs. In addition, the recent contributions in
\cite{deeproadmapper}, \cite{dlinknet},
\cite{batra2019improved}, \cite{singh2018self},
\cite{rotich2018using}, \cite{rotich2018resource} and
\cite{costea2018roadmap} use datasets with precise training
labels for semantic labeling of overhead imagery. All these
approaches use single-view images that are usually
pre-orthorectified.

Some examples of popular datasets for semantic labeling of
overhead imagery with manually-generated and/or
manually-corrected training labels can be found in
\cite{deepglobe}, \cite{spacenet}, \cite{jhu_us3d},
\cite{jhu_urban3d}, \cite{jhu_new3d}, \cite{datafusion},
\cite{datafusioncon1}, \cite{isprs2d}, \cite{torontocity}
and \cite{dota}. The dataset presented in
  \cite{jhu_us3d} provides satellite images, airborne LiDAR,
  and building labels (derived from LiDAR) that are manually
  corrected. The DeepGlobe dataset \cite{deepglobe} provides
  satellite images and precise labels (annotated by experts)
  for land cover classification and road and building
  detection. The study described in \cite{jhu_new3d}
  combines multi-view satellite imagery and large-area DSMs
  (obtained from commercial vendors) \cite{jhu_urban3d} with
  building labels that are initialized using LiDAR from the
  HSIP 133 cities data set \cite{GRID}. The IEEE GRSS Data
  Fusion Contest dataset \cite{datafusion},
  \cite{datafusioncon1} provides true ortho images, LiDAR
  and hyperspectral data along with precise groundtruth
  labels for 17 local climate zones. A summary of the
  top-performing algorithms on this dataset can be found in
  \cite{datafusioncon2}.

We will restrict our discussion of prior contributions that
use information from multiple views to CNN-based
approaches. Variants of multi-view CNNs have been proposed
primarily for segmentation of image-sequences and video
frames, and for applications such as 3D shape
recognition/segmentation and 3D pose
estimation. State-of-the-art examples include the approaches
described in \cite{mv-paper1}, \cite{mv-paper2},
\cite{mv-paper3}, \cite{mv-paper4}, \cite{mv-paper5},
\cite{mv-paper6}, \cite{mv-paper8}, \cite{mv-paper9},
\cite{mv-paper10}, \cite{mv-paper11}, \cite{mv-paper12},
\cite{mv-paper13} and \cite{mv-paper14}. These contributions
share one or more of the following attributes: (1) They
synthetically generate multiple views by either projecting
3D data into different planes, or by viewing the same image
at multiple scales; (2) They extract features from multiple
views, concatenate/pool such features and/or enforce
consistency checks between the features; (3) They use only a
few views (of the order of 5 or less). For instance,
  while the study described in \cite{mv-paper1} improves
  semantic segmentation of RGB-D video sequences by
  enforcing consistency checks after projecting the
  sequences into a reference view at training time, the
  approach presented in \cite{mv-paper3} estimates 3D hand
  pose by first projecting the input point clouds onto
  three planes, subsequently training CNNs for each plane
  and then fusing the output predictions.

With respect to the field of remote-sensing, multi-date
  satellite images have been used for applications such as
  change detection. For instance, the study described in
  \cite{changedetect} demonstrates unsupervised
  change-detection between a single pair of images with deep
  features extracted using a cycle-consistent GAN
  \cite{CycleGan2017}. However, there do not exist many studies
that use CNNs for labeling of multi-view and multi-date
satellite images. A relevant contribution is the one
described in \cite{dfc_winner} that won the 2019 IEEE GRSS
Data Fusion Contest for Multi-View Semantic Stereo
\cite{dfc2019}. The work in \cite{danesfield} also uses
off-nadir WV3 images for semantic labeling.  Both these
approaches still treat the different views of the same scene
on the ground independently during training. To the best of
our knowledge, there has not existed prior to our work reported here a true multi-view
approach for semantic segmentation using satellite images.

We also include a brief review of the literature related to
constructing DSMs from satellite images. Fully automated
approaches for constructing DSMs from satellite images have
been discussed in \cite{ohio_3d}, \cite{largescale_isprs},
\cite{nasa-ames}, \cite{ozge_comparison}, \cite{s2p}, \cite{demsculpt} and
\cite{isprs_ptcld_example}. While the studies described
  in \cite{largescale_isprs}, \cite{nasa-ames} and
  \cite{s2p} process pairs of images to construct multiple
  pairwise point clouds that are subsequently fused to
  construct a dense DSM, the contribution in
  \cite{ozge_comparison} compares such approaches with an
  alternative approach that divides the 3D scene into
  voxels, projects each voxel into all the images and
  subsequently reasons about the probability of occupancy of
  each voxel using the corresponding pixel features from all
  the images. {\em In all of these contributions, the DSMs
  that are constructed cover relatively small
  areas.
} The large-area DSM contribution in
\cite{Pleiades} is based on a small number of in-track
images that are typically captured seconds or minutes apart
by the Pl\'eiades satellite. In addition to the aforementioned  contributions, the study in \cite{rvlstereo} provides a dataset containing stereo-rectified images and the associated groundtruth disparity maps for different world regions, that can be used for benchmarking stereo-matching algorithms.

\section {A Framework for Large-Area Image
  Alignment, DSM Creation, and Generating Training Samples from OSM}
\label{sec:system_overview}

\begin{figure*}[h]
  \centering
  \includegraphics[width=0.7\linewidth]{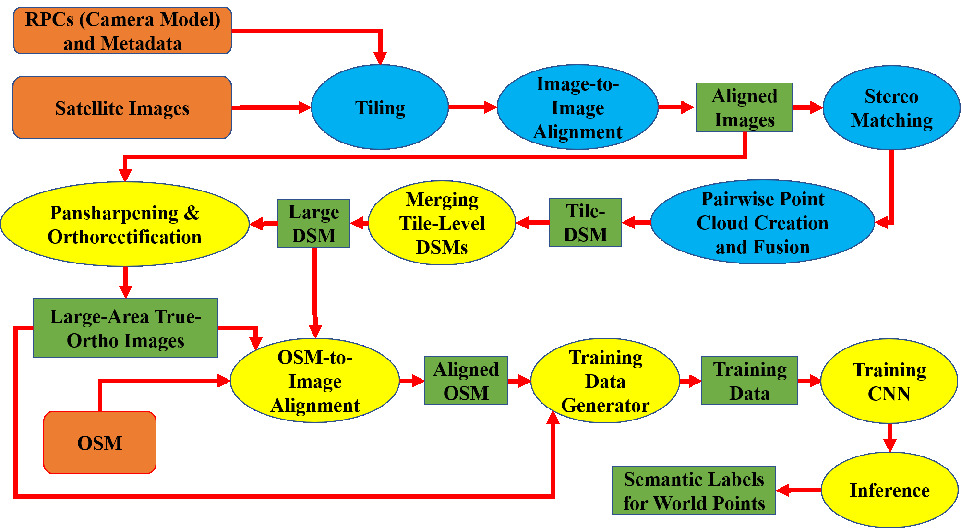}
  \caption{Overview of our framework. The three inputs
      are shown in orange-colored boxes. All outputs
      produced by the system are shown in green-colored
      boxes. The modules in blue-colored ellipses operate on
      a tile-wise basis.}
  \label{fig:overview}
\end{figure*}

As stated in the Introduction section, our goal is to generate
accurate semantic labels for the points on the ground (as
opposed to the pixels in the images). Solving this problem
requires correcting the positioning errors in the satellite
cameras and estimating accurate elevation information for
each point on the ground --- since only then we can
accurately establish the relationship between the pixels in
the images and the points on the ground. This will also
enable us to establish correspondences between the pixels of
the multiple views of the same scene.

Therefore, an important intermediate step in our processing
chain is the calculation of the DSM.  To the best of our
knowledge, there is no public contribution that discusses a
complete framework for automatic alignment and creation of
{\em large-area} DSMs over a 100 km$^2$ region using
satellite images taken as far apart as 2 years.  Because of
the role played by high-quality large-area DSMs in our
framework, we have highlighted this part of the framework in
the Introduction and shown some sample results in
Fig. \ref{fig:dsm_example}.

An overview of the overall framework presented in this study
is shown in Fig. \ref{fig:overview}. The system has three
inputs that are shown by the orange colored boxes: (1)
panchromatic and 8-band multispectral satellite
images;   
(2) the metadata associated with the images; and (3) the OSM
vectors.  After the CNN is trained in the manner described
in the rest of this manuscript, the framework directly outputs
semantic labels for the world points. 
In the rest of this section, we will briefly describe the
major components of the framework, apart from the
machine-learning component. These components are described
in greater detail in the Appendices.

\begin{LaTeXdescription}%
  \setlength\itemsep{2pt}
\item[Tiling and Image Alignment:] The notion of a tile is
  used only for aligning the images and for constructing a
  DSM. For the CNN-based machine-learning part of the
  system, we work directly with the whole images and with
  the OSM for the entire area of interest. Tiling is made
  necessary by the following two considerations: (1) The
  alignment correction parameters for a full satellite image
  cannot be assumed to be the same over the entire image;
  (2) The computational requirements for image-to-image
  alignment and DSM construction become too onerous for
  full-sized images. We have included evidence for the need
  for tiling in Appendix \ref{sec:image_align}. On a
    related note, the study reported in \cite{chipcluster}
    describes an approach that divides a large region into
    smaller chips for the purpose of land cover clustering.
  After tiling, the images are aligned with
  bundle-adjustment algorithms, which is a standard practice
  for satellite images. Alignment in this context means
  calculating corrections for the rational polynomial
  coefficients (RPCs) of each image.

\item[DSM Construction:] A DSM is constructed from the
  disparity map generated by the hierarchical tSGM algorithm
  \cite{rothermel2012sure}.  Stereo matching is only applied
  to those pairs that pass certain prespecified criteria
  with respect to differences in the view angles, sun
  angles, time of acquisition, etc., subject to the
  maximization of the azimuth angle coverage. The disparity
  maps and corrected RPCs are used to construct pairwise
  point clouds. Since the images have already been aligned,
  the corresponding point clouds are also aligned and can be
  fused without any further 3D alignment. Tile-level DSMs
  are merged into a large-area DSM.

\item[Generating Training Samples:] The training data is
  generated by using an $F \times F$ window to randomly
  sample the images after they have been pansharpened and
  orthorectified using the DSM.  We refer to such an $F \times F$
  window on the ground as a ground-window.
  The parameter $F$ is empirically set to 572 in our experiments\footnote{Note that the value of $F$ can change depending on the resolution of the images. $F$ should be chosen such that the windows are large enough to capture sufficient spatial context around the objects of interest.}. Subsequently, the OSM vectors are converted to raster format with
  the same resolution as in the orthorectified images. Thus
  there is a label for each geographic point in the
  orthorectified images. The OSM roads are thickened to have
  a constant width of 8m. Since the images
  are aligned with sub-pixel accuracy and are
  orthorectified, the training samples from the multiple
  images that view the same ground-window correspond to
  one another on a point-by-point basis, thereby giving us
  multi-view training data.
\end{LaTeXdescription}

\section{Multi-View Training and Inference}
\label{sec:mvapproach}
\subsection{Motivation for our Proposed Approach} 
\label{sec:motiv}

Our multi-view training framework is motivated by the
following factors:

\begin{LaTeXdescription}
  \setlength\itemsep{2pt}
\item[Convenience:] With newer and better single-view CNNs
being designed so frequently, it would be convenient if the
multi-view fusion module could be designed as an add-on to
an existing pretrained architecture. This would make it easy
to absorb the latest improvements in the single-view
architectures directly into the multi-view fusion
framework. We won't have to rethink the feature
concatenation for each new single-view CNN
architecture. Additionally, we want to efficiently train the
single-view weights in parallel across multiple GPUs and
carry out fusion on a single GPU.

\item[Multi-Date Images:] The satellite images could have
been collected years apart under different illumination and
atmospheric conditions.  Thus, our task is very different
from traditional multi-view approaches that work with 3D
shapes or images captured by moving a (handheld)
camera around the same
scene.  
\item[Varying Number of Views:] 
The number of views covering a ground-window can vary between
1 to all available images (32 in our case). This causes
practical challenges in backpropagating gradients 
when using CNNs that assume the availability of a fixed
number of views for concatenating features. At the same
time, we do not want to exclude windows that are covered by
less than a specified number of views. Our goal is to use
all available training data and all available views for
every ground-window.
\end{LaTeXdescription}
\subsection{Multi-View Fusion Module}

\label{sec:mvcnn}
\begin{figure*}[h]
  \centering
  \includegraphics[width=1\linewidth]{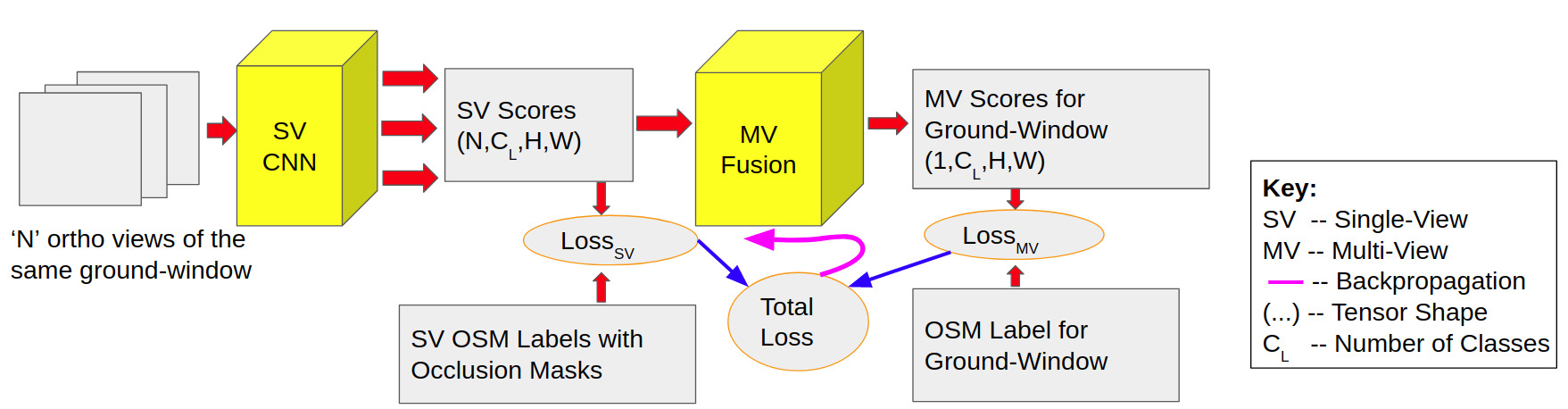}
  \caption{Overview of Multi-View Training}
  \label{fig:mvtrain}
\end{figure*}

\begin{figure*}[h]
  \centering
  \subfloat[]
  {
    \includegraphics[width=1\linewidth
    ]{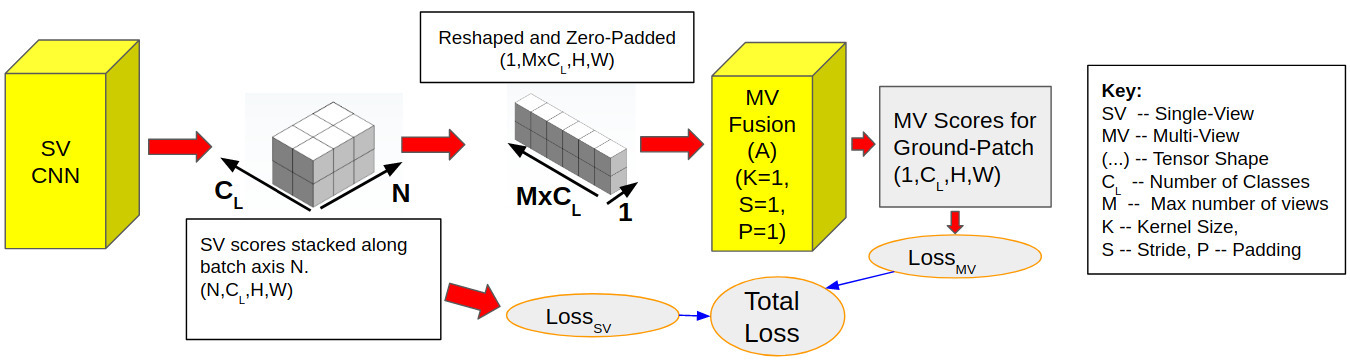}
  }\\
  \subfloat[]
  {
    \includegraphics[width=1\linewidth
    ]{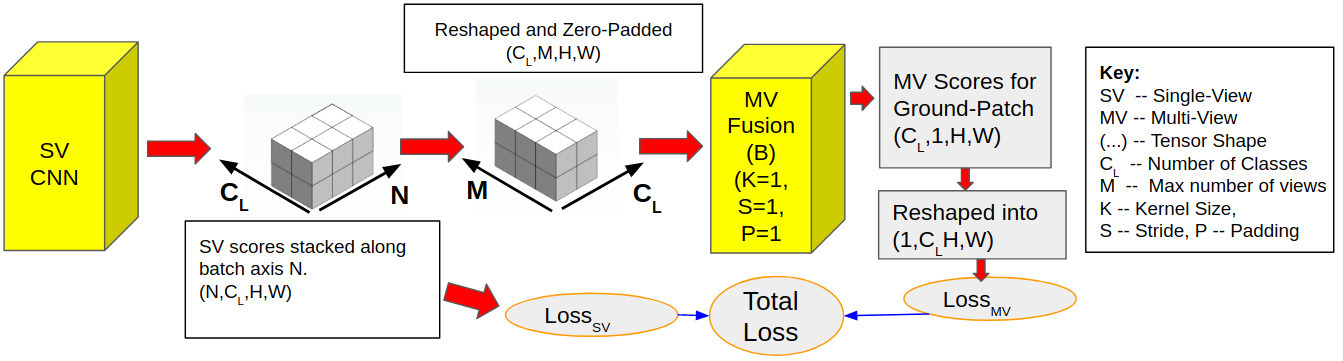}
  }
  \caption{Two choices for Multi-View Fusion. At top is MV-A
    in which the weights of the MV Fusion layer are
    different for each channel of each view. At bottom is
    MV-B where the weights of the MV Fusion layer are shared
    by all the channels of a view.}
  \label{fig:mv-ab}
\end{figure*}

Fig. \ref{fig:mvtrain} shows an overview of our multi-view
training framework where we propose that the multi-view
information be aggregated at the predictions stage. In this
sense, our approach is related to the strategies discussed
in \cite{mv-paper9} and \cite{mv-paper14}. While the
contribution in \cite{mv-paper9} considers the ``RGB'' and
the depth channel of the same RGB-D image as two ``views''
(which is a much simpler case), the 3D shape segmentation
approach in \cite{mv-paper14} synthetically generates
multiple-views of the same 3D object. In contrast, the
significantly more complex nature of our data makes our
problem very different from these
tasks.

The multi-view fusion module shown in Fig. \ref{fig:mvtrain}
can be added to any existing/pretrained single-view CNN. We
experimented with different choices for this module and
present two that gave good performance yields. These are
shown in Fig.~\ref{fig:mv-ab} and we denote them as MV-A
(Multi-View-A) and MV-B(Multi-View-B) respectively. Both
MV-A and MV-B consist of a single block of weights with
kernel size, stride and padding set to 1.

In the following discussion, $V$ denotes a subset of views
for a single ground-window. $N$ is the number of views in
$V$. $H$ and $W$ are the height and width of a single view
respectively. $M$ is the maximum number of possible views
for a ground-window. $C_L$ is the number of target classes.

As shown in Fig. \ref{fig:mvtrain}, the Single-View (SV) CNN
outputs a tensor of shape $(C_L,H,W)$ for each of N views
which are concatenated along the batch axis to yield a
tensor of shape $(N,C_L,H,W)$, which we denote as
$T^N_{MV}$. This tensor is then inserted into a larger tensor
which we denote as $T_{MV}$. Each view has a fixed index in
$T_{MV}$. Missing views are filled with zeros. The
difference between MV-A and MV-B can now be explained as
follows.

\begin{LaTeXdescription}
  \setlength\itemsep{2pt}
\item[MV-A:] In this case, $T^N_{MV}$ is reshaped into a tensor
of shape $(1,N \times C_L,H,W)$. It is then inserted into
$T_{MV}$ which is of shape $(1,M \times C_L,H,W)$. $T_{MV}$
is then input to the MV-A module which subsequently outputs
a tensor of shape $(1,C_L,H,W)$. MV-A thus contains a total
of $M \times C_L$ trainable weights, one for each channel of
each view.

\item[MV-B:] In this case, $T^N_{MV}$ is first reshaped into a
tensor of shape $(C_L,N,H,W)$. It is then inserted into
$T_{MV}$ which is of shape $(C_L,M,H,W)$. $T_{MV}$ is then
input to the MV-B module which subsequently outputs a tensor
of shape $(C_L,1,H,W)$.  MV-B thus contains a total of $M$
trainable weights, one for each view. The first and second
axis of this tensor are swapped to yield a tensor of shape
$(1,C_L,H,W)$ which is then used to calculate the loss.
\end{LaTeXdescription}

\subsection{Multi-View Loss Function}
\label{sec:loss}

The total loss is defined as 

\begin{align}
  L = \alpha \cdot L_{SV} + \beta \cdot L_{MV}
\label{eqn:loss}
\end{align}

where $L_{SV}$ represents the single-view loss, $L_{MV}$
represents the multi-view loss and $\alpha$ and $\beta$ are
scalars used to weight the two loss functions. The
single-view loss is calculated as follows.

\begin{align}
  L_{SV} = \frac{1}{N} \sum_{i=1}^{N} \text{CE$_i(G_i,T_i)$}
\end{align}

where CE$_i$ is the pointwise cross-entropy loss for the
$i^{th}$ view, $N$ is the number of views in a subset $V$
of views that cover a single ground-window and $T_i$ is the
output tensor of the SV CNN for the $i^{th}$ view. To
calculate CE$_i$, we mask the OSM labels for the
ground-window with the occlusion mask of the $i^{th}$
view. This masked ground-truth is denoted by $G_i$ in the
equation above. Note that this mask is implicitly computed
during the process of true orthorectification. The gradients
of $L_{SV}$ are not backpropagated at these masked
points. What this means is that for each individual view,
$L_{SV}$ only focuses on portions of the ground-window that
are visible in that view.

The pointwise cross-entropy loss between two probability
distributions $A$ and $B$, each defined over $C_L$ classes,
is calculated as follows.

\begin{align}
  \text{CE}(A,B) = -\sum_p\sum^{C_L}_{j=1} A(p,j) \cdot log(B(p,j))
  \label{eqn:celoss}
\end{align}

where $p$ refers to a single point. $A(p,j)$ is the
probability that point $p$ belongs to class $j$ as defined
by $A$. $B(p,j)$ is the probability that point $p$ belongs
to class $j$ as defined by $B$.

The multi-view loss is calculated as follows.
\begin{align}
  L_{MV} = \text{CE}(G,P_{MV})
\label{eqn:mvloss}
\end{align}

where CE$(G,P_{MV})$ is the pointwise cross-entropy loss for
the ground-window. This is calculated using the unmasked OSM
label $G$ and the output $P_{MV}$ of the MV Fusion
module. $P_{MV}$ can be viewed as a final probability
distribution that is estimated by fusing the individual
probability distributions that are output by the SV CNN for
each of the $N$ views. We can denote $P_{MV}$ as a function
$f(T_1,T_2,...,T_N)$ where $f$ depends upon the architecture
of the MV Fusion module. Note that $f$ is differentiable.
Thus, Eq. \ref{eqn:mvloss} can be rewritten as

\begin{align}
  L_{MV} = \text{CE}(G,f(T_1,T_2,...,T_N))
\label{eqn:mvloss1}
\end{align}

Substituting the expression for the $\text{CE}$ loss from
Eq. \ref{eqn:celoss} into Eq. \ref{eqn:mvloss1}, we get the
following expression for $L_{MV}$.

\begin{align}
  L_{MV} = -\sum_p\sum^{C_L}_{j=1} G(p,j) \cdot log(f(T_1,T_2,...,T_N)(p,j))
\label{eqn:mvloss2}
\end{align}

Note that \textit{$L_{MV}$ is not linearly separable over
  the views in V.} In other words, unlike $L_{SV}$, we
cannot separate it into a sum of losses for each view. Thus,
$L_{MV}$ captures the predictions of the network in an
ensemble sense over multiple views covering a
ground-window. When backpropagating the gradients of $L$,
the gradients from $L_{MV}$ are influenced by the relative
differences between the predictions for each view, and this
in turn translates into better weight-updates. Moreover, by
using $L_{MV}$, the network is shown labels for all portions
of the ground including those that are missing in some views
of $V$. This enables the network to make better decisions
about occluded regions using multiple views.

\subsection{Strategies for Multi-View Training and Inference}
\label{sec:mvtrain}

\subsubsection{Approaches for Data-Loading}

The term ``data-loading'' refers to how the data samples are
grouped into batches and input to the CNN. We use two
different data-loading approaches.

\begin{LaTeXdescription}
  \setlength\itemsep{2pt}
\item[Single-View Data-Loading (SV DATALOAD):] This is a
  conventional data-loading strategy where a single training
  batch can contain views of different ground-windows. The
  batch size is constant and only depends on the available
  GPU memory. SV DATALOAD uses all the available data.

\item[Multi-View Data-Loading (MV DATALOAD):] 
  Under this strategy, a training batch consists solely of
  views that cover the same ground-window. The number of
  such views can vary from window to window. However, due to
  memory constraints, we cannot load all 32 views onto the
  GPUs simultaneously. As a work around, we use the
  following approach. Let $|Q|$ denote a pre-specified
  number of views that can fit into the GPU memory, $R$
  denote the set of available views for a ground-window and
  $|R|$ denote the total number of views in $R$. If
  $|R| < |Q|$, we skip loading this ground-window. If
  $|R| > |Q|$, we randomly split $R$ into a collection of
  overlapping subsets $\{Q_j\}$, such that each $Q_j$ has
  $|Q|$ views and $\cup Q_j = R$ where $\cup$ denotes the
  union operator. The tensor $T_{MV}$ that is input to the
  MV Fusion module is reset to zero before inputting each
  $Q_j$ to the CNN. Note that \textit{this random split has
    the added advantage that the CNN sees a different
    collection of views for the same ground-window in
    different epochs, which should help it to learn better.}
\end{LaTeXdescription}

The design of MV DATALOAD is motivated by our
  observation that if we allow the batch size to change significantly
  for every ground-window (based on the corresponding number
  of available views), it significantly slows down the rate
  of convergence. Therefore, we exclude ground-windows with
less than $|Q|$ views, and for the remaining windows we make
sure that every subset $Q_j$ has $|Q|$ views. \textit{This
  enforces a constant batch size of $|Q|$}, resulting in
faster convergence.

\subsubsection{Training Strategies}

We use the following different strategies to train the CNNs.
\\\\
\noindent \textbf{Single-View Training (SV TRAIN):} In this strategy,
the SV CNN is trained independently of the MV Fusion
module. We apply the SV DATALOAD approach to use all
available data. One can also interpret this as setting
$\beta = 0$ in Eq. \ref{eqn:loss} and freezing the weights
of the MV Fusion module.

We now define three different multi-view training strategies
as follows.

\begin{LaTeXdescription}%
  \setlength\itemsep{2pt}
\item[MV TRAIN-I:] We first train the SV CNN using SV
  TRAIN. Subsequently, we use MV DATALOAD to only train the
  MV Fusion module by setting $\alpha=0$ in
  Eq. \ref{eqn:loss}, and by freezing the weights of the SV
  CNN. Hence, $L_{MV}$ only affects the weights of the MV
  Fusion module and does not affect the SV CNN.

\item[MV TRAIN-II:] We first train the SV CNN using SV
  TRAIN. Subsequently, both the pretrained SV CNN and the MV
  Fusion module are trained together using MV DATALOAD and
  the total loss as defined in Eq. \ref{eqn:loss}. Thus, the
  $L_{MV}$ loss influences the weight updates of the SV CNN
  as well. In practice, we lower the initial learning rate
  of the SV CNN as it has already been trained and we only
  want to fine-tune its weights.

\item[MV TRAIN-III:] In this strategy, we do not pretrain
  the SV CNN, but rather train both the SV CNN and the MV
  Fusion module together from scratch using the total loss
  $L$ (Eq. \ref{eqn:loss}), and MV DATALOAD. This has the
  disadvantage that the network never sees ground-windows
  with less than $|Q|$ views, where $|Q|$ is a
  user-specified parameter. One might expect this reduction
  in the amount of training data to negatively impact
  performance, especially given the sparse nature of the OSM
  labels. Our experimental evaluation confirms this.
\end{LaTeXdescription}

To make a decision on when to stop training, a common
practice in machine-learning is to use a validation
dataset. However, in our case the validation data is also
drawn from OSM (to avoid any human intervention), and is
therefore noisy. To handle this, we make the following
proposal. We train a network until the training loss stops
decreasing. At the end of every epoch, we measure the IoU
using the validation data. For inference, we save the
network weights from two epochs -- one with the smallest
validation loss and the largest validation IoU, and the
other with the smallest training loss and an IoU that is
within an acceptable range of the largest validation IoU (to
reduce the chances of overfitting to the training data). We
denote the former as EPOCH-MIN-VAL (E$_{\text{MIN-VAL}}$)
and the latter as EPOCH-MIN-TRAIN (E$_{\text{MIN-TRAIN}}$)
respectively.

\subsubsection{Inference}

To establish a baseline, we use a SV CNN trained with the SV
TRAIN strategy defined above, and merge the predictions from
overlapping views via majority voting. We will denote this
approach as SV CNN + VOTE. We also implemented an
alternative strategy of simply averaging the predicted
probabilities across overlapping views, which produced
nearly identical results to majority voting. For the sake of
brevity, we only report the results from SV CNN + VOTE as
the baseline.

Inference using the SV CNN + MV Fusion module is noticeably
faster than SV CNN + VOTE, because the former combines
multi-view information directly on the GPU.  For inference,
the MV DATALOAD approach can be used with a single minor
modification. Instead of resetting the $T_{MV}$ tensor to
zeros before inputting each subset $Q_j$ of $R$, it is only
reset to zeros once for each ground-window. This means that the
final prediction for a ground-window is still made using all
the views.

\section{Semantic Segmentation Using Off-Nadir Images}
\label{sec:offnadir}

Up till now, our discussion was focused on using true
orthophotos for semantic segmentation. However, for many
applications, it would be useful to directly train CNNs on
the off-nadir images. Even for labeling world points, it
would be interesting to compare the approach from the
previous section vis-a-vis first training CNNs on the
original off-nadir images, and subsequently orthorectifying
the predicted labels. However, this would require a way to
project the OSM training labels from geographic coordinates
into the off-nadir images. Most prior OSM-based studies in
the literature are ill-equipped to carry out such a
comparison because they use pre-orthorectified images. Our
end-to-end automated pipeline, which includes the ability to
create large-area DSMs, enables us to solve the problem
stated above in the manner described below.

Since each building and buffered road-segment is represented
by a polygon in OSM, we use the following procedure to
create smooth labels in the off-nadir images. For a specific
polygon and a specific off-nadir image,
\begin{enumerate}
\item We obtain the longitude and latitude coordinates of
  the vertices of the polygon from OSM.
\item Using the longitude and latitude coordinates, we find
  the corresponding height values of the vertices from the
  DSM.
\item Using the RPC equations and the latitude, longitude
  and height coordinates, we project each vertex into the
  off-nadir image.
\item Subsequently all the pixels contained inside a
  projected polygon are marked with the correct
  label. Portions of the polygon that fall outside the image
  are ignored.
\end{enumerate}

The above procedure is repeated for every polygon and
off-nadir image. In practice, the polygons can be projected
independently of one another in parallel. This method is
very fast, but does come at a cost. Consider an example of
projecting a polygon representing a building-roof into an
off-nadir image. If the DSM height for a corner of this
polygon is incorrect, then, because we first project vector
data into the image and subsequently rasterize it, the
projected shape of the entire building-roof label could
become distorted. \textit{Thus, the noise in the DSM has
  greater impact on the noise in the training labels when
  using off-nadir images vis-a-vis using true orthophotos.}

A possible alternative strategy is to first map each pixel
in each off-nadir image into its longitude and latitude
coordinates, and subsequently check if this point lies
inside an OSM polygon. However, inverse projection needs an
iterative solution and cannot be done directly with the RPC
equations. Such a strategy will be significantly slower than
our adopted method.

\section{Experimental Evaluation and Results}
\label{sec:evaluation}
We use two datasets to evaluate the different
components of our framework. The first dataset consists of
32 WV3 images covering a 120 km$^2$ region in Ohio and the
second dataset consists of 32 WV3 images covering a 62
km$^2$ region in California. The latter is part of the
publicly available Spacenet \cite{spacenet}
repository. Building and road label data is downloaded from
the OSM website. \emph{No other preprocessing is done before
  feeding the data to our framework}. Alignment and
large-area DSM construction are evaluated using both
datasets. For an extensive quantitative assessment of the
performances of the different semantic segmentation
strategies, we divided the 120 km$^2$ region in Ohio into a
109 km$^2$ region for training, a 1 km$^2$ region for
validation, and an unseen 10 km$^2$ region for
inference. The unseen region contains precise manual
annotations.
The last region is ``unseen'' because no samples in the
training and the validation regions fall inside that region.

We select the popular U-Net \cite{unet} as the SV CNN
because it is lightweight and has been used in many prior
studies with overhead imagery \cite{resunet},
\cite{yang2019road}, \cite{yang2019road_v2}. The U-Net is
modified to accept 8 band data, and we add
batch-normalization \cite{batchnorm} layers. Since OSM
labels are sparse, we weight the cross entropy losses with
the weights set to 0.2, 0.4 and 0.4 for the background,
building and road classes respectively. Training is done
using 4 NVIDIA Gtx-1080 Ti GPUs. Due to GPU memory
constraints, the parameter $|Q|$ for MV DATALOAD is set to
16.

We will present the results of the semantic-segmentation
studies in the main body of the manuscript. Quantitative
evaluation of the image-to-image alignment and inter-tile
DSM alignment are included in Appendix
\ref{sec:align_qual}.

\subsection{Single-View vs Multi-View CNNs}

We have carried out experiments with different combinations
of CNNs, training strategies and inference
models. 
For clarity, we present the most interesting results in this
manuscript. 
The relevant notations have already been
defined in Section \ref{sec:mvtrain}. 
To assist the reader, we will explain the notation used in
the tables below with an example. Consider the first row in
Table \ref{table:sv-mv}. This row corresponds to the case of
training a Single-View CNN using SV TRAIN. At inference
time, the EPOCH-MIN-VAL weights are used and the predictions
from different views are merged using majority voting.

\begin{table}[h]
  \renewcommand{\arraystretch}{1.5}
  \caption{Comparison of SV TRAIN vs MV TRAIN-II}
  \label{table:sv-mv}
  \centering
  \begin{tabular}{|p{2.05cm}|p{1.675cm}|p{1.175cm}|p{1cm}|p{0.6cm}|}
    \hline
    CNN & Training & Inference & \multicolumn{2}{|c|}{IoU}\\
    \hline
        & & & Buildings & Roads\\
    \hline
    SV CNN + VOTE & SV TRAIN & E$_{\text{MIN-VAL}}$ & 0.75 & 0.57 \\
    \hline
    SV CNN + MV-A & MV TRAIN-II & E$_{\text{MIN-VAL}}$ & \textbf{0.79} & 0.55 \\
    \hline
    SV CNN + MV-B & MV TRAIN-II & E$_{\text{MIN-VAL}}$ & \textbf{0.80} & 0.57\\
    \hline
    \hline
    SV CNN + VOTE & SV TRAIN & E$_{\text{MIN-TRAIN}}$ & 0.75 & 0.56 \\
    \hline
    SV CNN + MV-A & MV TRAIN-II & E$_{\text{MIN-TRAIN}}$ & 0.73 & \textbf{0.6}\\
    \hline
    SV CNN + MV-B & MV TRAIN-II & E$_{\text{MIN-TRAIN}}$ & 0.73 & \textbf{0.64}\\
    \hline
  \end{tabular}
\end{table}

Table \ref{table:sv-mv} shows the best gains that we get by
using multi-view training and inference, vis-a-vis
single-view training and majority voting. The first three
rows correspond to running inference using the EPOCH-MIN-VAL
weights. Using MV TRAIN-II to train the SV CNN + MV-B
network, we outperform the baseline with a $5\%$ increase in
the IoU for the building class, while performing comparably
with the baseline for the road class. With the MV-A module,
the IoU for the building class improves by $4\%$, but that of
the road class decreases by $2\%$.

The noise in the training and validation labels for roads is
much more than that for buildings because we assume a
constant width of 8 m for all roads, and because the
centerlines of roads (as marked in OSM) are often not along
their true centers. To handle this, in Section
\ref{sec:mvtrain}, we proposed to also save the network
weights for the epoch with the minimum training loss and
good validation IoU. By using the validation IoU, we
reduce the chances of these network weights being overfitted
to the data. Our intuition is borne out by the last three
rows of Table \ref{table:sv-mv}. When compared to the
baseline, using MV TRAIN-II with the SV CNN + MV-A and the
SV CNN + MV-B networks increases the IoU for the road class
by $4\%$ and $8\%$ respectively while slightly lowering the
building IoU by $2\%$. It is interesting to note that in
contrast, EPOCH-MIN-VAL and EPOCH-MIN-TRAIN perform
comparably for the SV TRAIN strategy. Based on these
results, we conclude that \textit{the MV TRAIN-II strategy
  is a good approach for multi-view training and the MV-B
  Fusion module yields the maximum gains.} We recommend
using EPOCH-MIN-VAL for segmenting buildings, and
EPOCH-MIN-TRAIN for segmenting roads.
We should point out that the SV CNN + VOTE baseline is trained on the
same data as the SV CNN + MV Fusion module (trained with MV
TRAIN II), and therefore the improvements are not due to data
augmentation.

\subsection{Does Multi-View Training Improve the Single-View
  CNN?}

To obtain additional insights into how multi-view training
improves accuracy, we carry out two ablation studies using
the SV CNN + MV-B network because it yielded the maximum
gains with the MV TRAIN-II strategy.

For the first study, we freeze the pretrained SV CNN and
only train the MV-B module using the MV TRAIN-I
strategy. The corresponding IoU scores are reported in the
first two rows of Table \ref{table:mv-I-ablation}. Comparing
these two rows with the baseline (SV CNN + VOTE) shown in
Table \ref{table:sv-mv}, we see that we do not get any
noticeable improvements. Remember that in MV TRAIN-I, the
multi-view loss ($L_{MV}$) only modifies the weights of the
MV Fusion module. This points to the need for allowing
$L_{MV}$ to influence the weights of the SV CNN as well, as
is done by MV TRAIN-II.

\begin{table}[h]
  \begin{center}
    \caption{Impact of multi-view training on the
      single-view CNN}
    \label{table:mv-I-ablation}
    \renewcommand{\arraystretch}{1.5}
    \begin{tabular}{|p{2.05cm}|p{1.675cm}|p{1.175cm}|p{1cm}|p{0.6cm}|}
      \hline
      CNN & Training & Inference & \multicolumn{2}{|c|}{IoU}\\
      \hline
          & & & Buildings & Roads\\
      \hline
      SV CNN + MV-B & MV TRAIN-I & E$_{\text{MIN-VAL}}$ & 0.75 & 0.57 \\
      \hline
      SV CNN + MV-B & MV TRAIN-I & E$_{\text{MIN-TRAIN}}$ & 0.75 & 0.57 \\
      \hline
      SV$_{\text{(MV)}}$ + VOTE & MV TRAIN-II & E$_{\text{MIN-VAL}}$ & \textbf{0.80} & 0.55 \\
      \hline
      SV$_{\text{(MV)}}$ + VOTE & MV TRAIN-II & E$_{\text{MIN-TRAIN}}$ & 0.74 & \textbf{0.62} \\
      \hline
      SV CNN + MV-B & MV TRAIN-II & E$_{\text{MIN-VAL}}$ & \textbf{0.80} & 0.57\\
      \hline
      SV CNN + MV-B & MV TRAIN-II & E$_{\text{MIN-TRAIN}}$ & 0.73 & \textbf{0.64}\\
      \hline
    \end{tabular}
  \end{center}
\end{table}

For the second study, we take the best performing SV CNN +
MV-B network that was trained using the MV TRAIN-II strategy
and remove the MV-B module from it. We denote this SV CNN as
SV$_{\text{(MV)}}$ CNN. We run inference using this
SV$_{\text{(MV)}}$ CNN and merge the predictions from
overlapping views using majority voting. The corresponding
IoUs are shown in the third and fourth rows of Table
\ref{table:mv-I-ablation}. Comparing these two rows with the
baseline SV CNN + VOTE in Table \ref{table:sv-mv}, we see
that \textbf {multi-view training has significantly improved
  the performance of the SV$_{\text{(MV)}}$ network itself,
  without any increase in the number of trainable
  parameters.} This indicates that intelligently training a
SV CNN using all the available views for a scene can
alleviate the effect of noise in the training labels,
without changing the original architecture of the SV
CNN. 
We reproduce the IoUs of the complete SV CNN + MV-B network
trained with MV TRAIN-II, in the fifth and sixth rows of
Table \ref{table:mv-I-ablation}. Comparing the 3$^{rd}$ and
5$^{th}$ rows, and the 4$^{th}$ and 6$^{th}$ rows, we see
that the MV Fusion module does provide an additional $2\%$
improvement in the IoU for the road class, over the
SV$_{\text{(MV)}}$ network.

\subsection{The Need for Using a Combination of SV DATALOAD and MV DATALOAD}

As another experiment, when we employ the MV TRAIN-III
strategy to train the SV CNN + MV-B network from scratch
using the MV DATALOAD method, the IoU for the building class
drops down significantly to 0.62, when compared to the
baseline in Table \ref{table:sv-mv}.  This is as expected
because in this case, the network is trained with fewer
training samples. It never sees ground-windows with less
than $|Q|$ views. Therefore, it is important that the
network be trained with as much non-redundant data as
possible and with multi-view constraints, as is done by
using a combination of SV DATALOAD and MV DATALOAD in MV
TRAIN-II.

\subsection{Comparison to Prior State-of-the-Art}
\label{sec:prior_comp}

For a fair comparison, we consider the most relevant prior
state-of-the-art studies that use multi-view off-nadir
images for semantic segmentation. The work presented in
\cite{dfc_winner} discusses the entry that won the 2019 IEEE
GRSS Data Fusion Contest for Multi-view Semantic
Stereo. This approach trains single-view networks using both
WV3 images and DSMs over a small 10-20 km$^2$ region with
\textit{ precisely annotated human labels} and reports an
IoU of about 0.8 for the building class. The performance gains
come from training the network on DSMs which helps to segment
buildings more accurately. Our best IoU for the building
class is also 0.8, but we use only noisy training labels
that are automatically derived from a much larger 100 km$^2$
region. It is possible that by adding the DSMs as inputs to
our network, we could further improve the
IoU.

Our IoU for the building class is noticeably better than
that reported by the work in \cite{danesfield}, which trains
single-view CNNs on WV3 images and OSM labels covering 1-2
km$^2$. 
Most of the other studies in the
literature use single-view pre-orthorectified images. It
should be pointed out that our multi-view training strategy
could be applied to any of those network architectures.
\hfill\\
\\
\\
\noindent \textit{Using DeepLabv3+ as the SV CNN:}
\\
\\
\indent For another comparison, we change the SV CNN from a
U-Net to a pretrained DeepLabv3+ (DLabv3) CNN with a
WideResNet38 trunk \cite{deeplabv3} that is one of the top
performers on the CityScapes benchmark dataset
\cite{Cordts2016Cityscapes}. We modify the first layer to
accept 8 bands. With this modification, the network has
$\sim$137 million trainable parameters whereas in comparison
the U-Net only has $\sim$31 million parameters.

We first train the network using SV TRAIN. 
Due to the size of the network, we set the batch size to 12
for SV TRAIN. At inference time we use majority voting. This
strategy is denoted as SV$_{(\text{DLabv3})}$ + VOTE. For
the multi-view training, we append the MV-B Fusion module to
the network and then employ the MV TRAIN-II strategy. Recall
that this is the best performing strategy for the
U-Net. $|Q|$ is set to 12 for the MV TRAIN-II strategy.

In Table \ref{table:deeplab-mv-train-II} we show the IoUs
for the DeepLabv3+ experiments. Firstly, we note that
SV$_{(\text{DLabv3})}$ + VOTE achieves an IoU of $0.828$ and
$0.553$ for the building and road classes respectively. It
should be kept in mind that this network has already been
trained on a large amount of precise labels from the
CityScapes dataset \cite{Cordts2016Cityscapes}.
What is interesting is that these numbers are comparable to
the corresponding IoUs of $0.80$ and $0.57$ for the U-Net +
MV-B network trained with MV TRAIN-II, despite the fact that
\textit{the U-Net has significantly fewer trainable
  parameters than the DeepLabv3+ network and that it has
  been trained only on noisy labels}.

The second row of Table \ref{table:deeplab-mv-train-II} has
the entry for running inference using the EPOCH-MIN-VAL
weights of the SV$_{(\text{DLabv3})}$ + MV-B network after
being trained with MV TRAIN-II. Compared to the
SV$_{(\text{DLabv3})}$ + VOTE, the building IoU goes down by
2.8$\%$ whereas the road IoU goes up by 5$\%$. The mean IoU
goes up by $1\%$ when we use multi-view training. One
possible reason for this small improvement is that we are
already at the limits of how much a CNN can learn, given the
extent of noise in the system. Another possibility is that
since the DeepLabv3+ network is much bigger than the U-Net,
and since the multi-view features are fused at the end, one
can expect the influence of the multi-view loss ($L_{MV}$)
on the earlier layers of the DeepLabv3+ network to be
reduced when compared to the case of the U-Net. We might get
better results by fusing the multi-view data at an earlier
stage in the network. This needs further
investigation.

\begin{table}[h]
\begin{center}
  \caption{Comparison of SV TRAIN with MV TRAIN-II when Using
    DeepLabv3+ as the SV CNN}
  \label{table:deeplab-mv-train-II}
   \renewcommand{\arraystretch}{1.5}
  \begin{tabular}{|p{2.3cm}|p{1.675cm}|p{1cm}|p{1cm}|p{0.6cm}|}
    \hline
    CNN & Training & Inference Model & \multicolumn{2}{c|}{IoU}\\
    \hline
        & & & Buildings & Roads\\
    \hline
    SV$_{(\text{DLabv3})}$ + VOTE  & SV TRAIN & E$_{\text{MIN-VAL}}$ & 0.828 & 0.553 \\
    \hline
    SV$_{(\text{DLabv3})}$ + MV-B & MV TRAIN-II & E$_{\text{MIN-VAL}}$ &  0.800 & 0.605 \\
    \hline
  \end{tabular}
\end{center}
\end{table}

\subsection{Training on True Orthophotos vs on Off-Nadir Images}

In Section \ref{sec:offnadir}, we have described our
framework for creating training labels to directly train a
SV CNN on the off-nadir images. For evaluation, we use this
trained CNN to label those portions of the off-nadir images
that correspond to the ``unseen'' inference region. For a
fair comparison, the predicted labels are then
orthorectified so that the evaluation is done in the same
orthorectified space. Predictions from overlapping images
are merged via majority voting.

When the off-nadir images and projected OSM labels are used
for training, both the EPOCH-MIN-VAL and EPOCH-MIN-TRAIN
weights yield IoU scores of 0.73 and 0.55 for the building
and road classes respectively. These scores are $2\%$ lower
than the corresponding numbers for the SV CNN + VOTE that is
trained on true orthophotos. As mentioned in Section
\ref{sec:offnadir}, one possible reason for this reduced IoU
might be the increased error in the OSM labels when
projected into the off-nadir images. Another reason could be
that the CNN finds it difficult to separate the building
walls from the roofs in the off-nadir images. In contrast,
vertical building walls are not present in true
orthophotos. Nevertheless, we have demonstrated that it is
possible to train a CNN on off-nadir images using noisy
labels, and obtain decent IoU scores. Such a CNN can be
directly used with new off-nadir images without having to
align or orthorectify the images. Multi-view training using
off-nadir images is also possible, albeit more challenging,
which we leave for future work.

\def \varwidth {0.4\linewidth}
\def \varheight {0.4\linewidth}
\begin{figure*}[h]
  \centering
  \subfloat[Ortho View]
  {
    \includegraphics[width=\varwidth, height = \varheight]{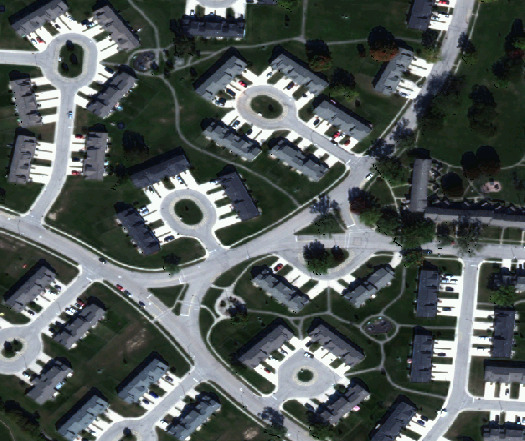}
  }
  \subfloat[Predicted Labels]
  {
    \includegraphics[width=\varwidth, height = \varheight]{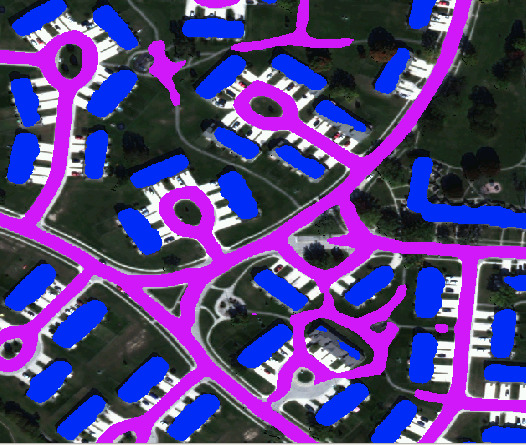}
  }\\
  \subfloat[Ortho View ]
  {
    \includegraphics[width=\varwidth, height = \varheight]{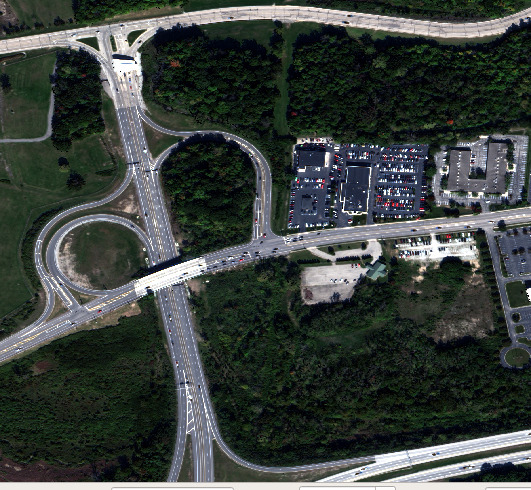}
  }
  \subfloat[Predicted Labels]
  {
    \includegraphics[width=\varwidth, height = \varheight]{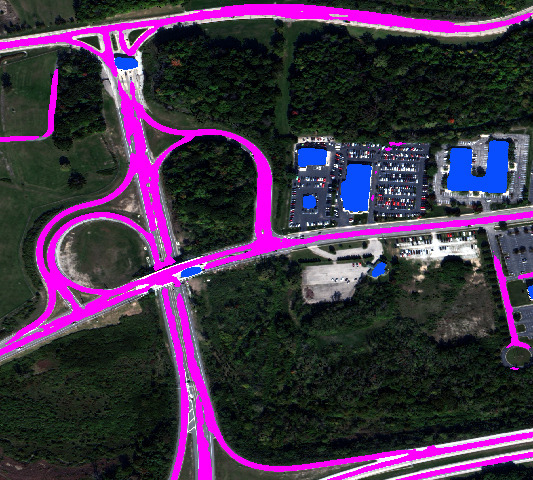}
  }\\
  \caption{Examples of orthorectified images and semantic
    labels output by our pipeline. Buildings are marked in
    \textcolor{blue}{blue} and roads are marked in
    \textcolor{magenta}{magenta}.}
  \label{fig:qual_eval_1}
\end{figure*}

\subsection{Qualitative Results}

\begin{figure*}[htbp!]
  \centering
  \subfloat[Ortho View]
  {
    \includegraphics[width=\varwidth, height = \varheight]{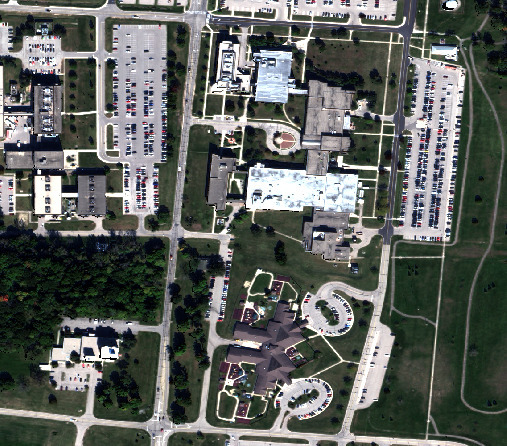}
  }
  \subfloat[Predicted Labels]
  {
    \includegraphics[width=\varwidth, height = \varheight]{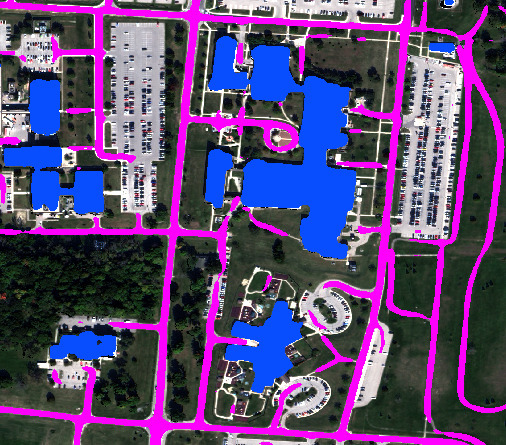}
  }\\
  \subfloat[Ortho View ]
  {
    \includegraphics[width=\varwidth, height = \varheight]{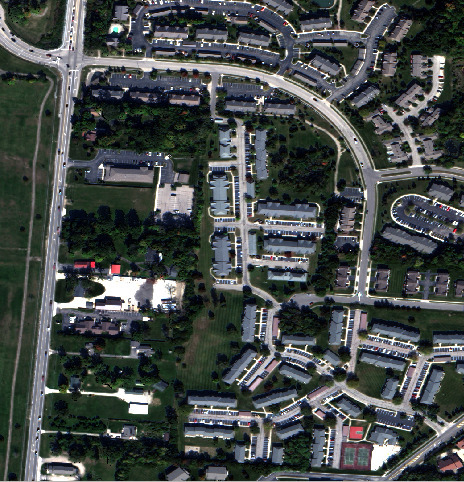}
  }
  \subfloat[Predicted Labels]
  {
    \includegraphics[width=\varwidth, height = \varheight]{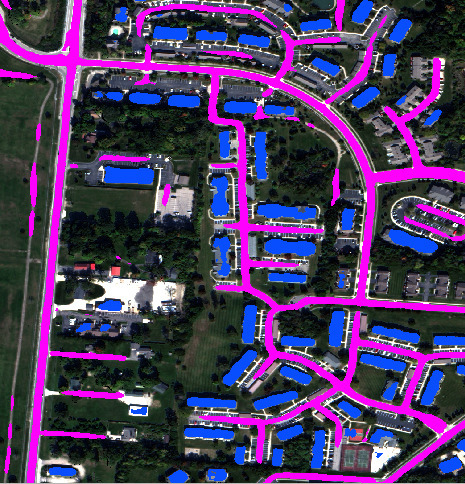}
  }\\
  \caption{Examples of orthorectified images and semantic
    labels output by our pipeline. Buildings are marked in
    \textcolor{blue}{blue} and roads are marked in
    \textcolor{magenta}{magenta}.}
  \label{fig:qual_eval_2}
\end{figure*}

In Figs. \ref{fig:qual_eval_1}, \ref{fig:qual_eval_2} and
\ref{fig:qual_eval_3}, we show some typical examples of
semantic labels output by our CNN. In addition, Fig.
\ref{fig:building-sv-mv} highlights how multi-view training
can help the CNN to segment challenging buildings such as
residential buildings which are often occluded by trees,
roofs made of highly reflective surfaces, and small
buildings. With respect to segmentation of roads, parking
lots pose a difficult challenge because their shapes and
spectral signatures are very similar to those of true
roads. However, multi-view training is able to learn from
the differences caused by the absence and presence of
vehicles in images captured on different dates, and this is
illustrated in Fig. \ref{fig:road-sv-mv}.

\begin{table*}[h]
  \setlength{\tabcolsep}{0.07cm}
  \begin{center}
    \begin{tabular}{|M{1.5cm}|M{0.25\linewidth}|M{0.25\linewidth}|M{0.25\linewidth}|}
      \hline
      Single-View Training & \raisebox{-0.5\height}{\includegraphics[width=1\linewidth, height = 0.8\linewidth]{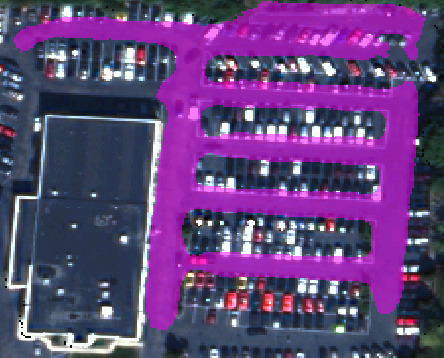}} 
      & \raisebox{-.5\height}{\includegraphics[width=1\linewidth, height = 0.8\linewidth]{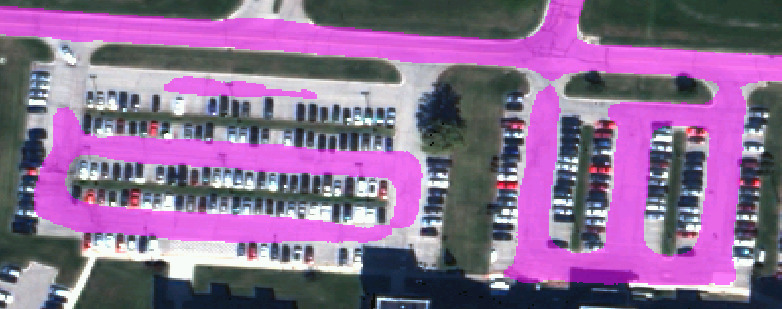}}
      & \raisebox{-.5\height}{\includegraphics[width=1\linewidth, height = 0.8\linewidth]{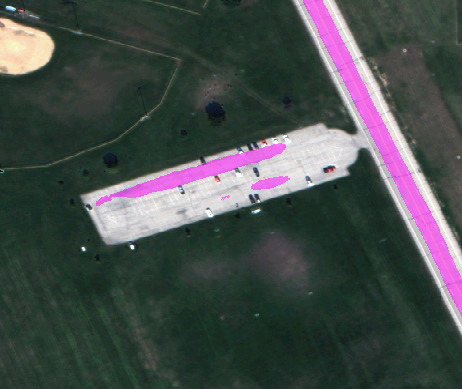}}\\
      \hline
      \hline
      Multi-View Training & \raisebox{-.5\height}{\includegraphics[width=1\linewidth, height = 0.8\linewidth]{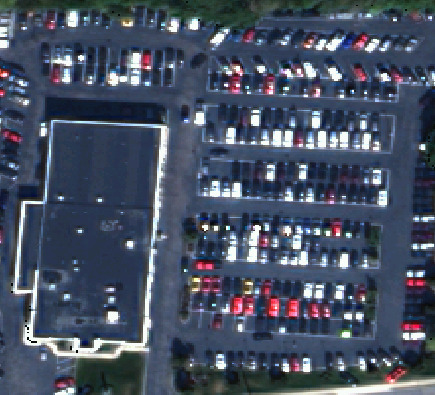}} 
      & \raisebox{-.5\height}{\includegraphics[width=1\linewidth, height = 0.8\linewidth]{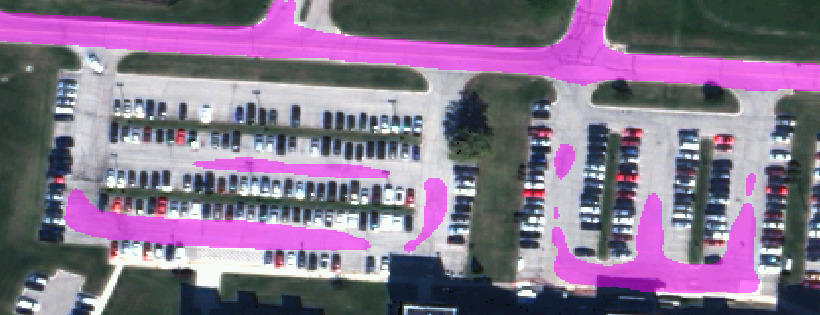}}
      & \raisebox{-.5\height}{\includegraphics[width=1\linewidth, height = 0.8\linewidth]{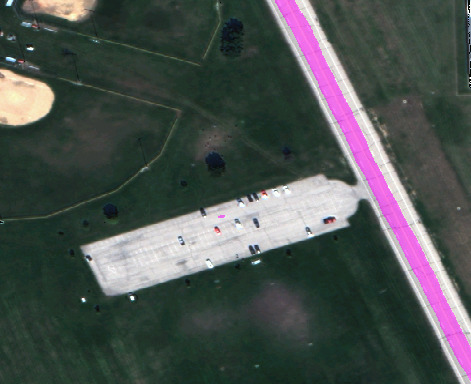}}\\
      \hline
    \end{tabular}
  \end{center}
  \captionof{figure}{Examples illustrating how multi-view
    training helps to distinguish parking lots from true
    roads. Predicted road labels are marked in
    \textcolor{magenta}{magenta}}
  \label{fig:road-sv-mv}
\end{table*}

\renewcommand \varwidth {0.4\linewidth}
\renewcommand \varheight {0.25\linewidth}

\begin{figure*}[htbp!]
  \centering
  \subfloat[Ortho View]
  {
    \includegraphics[width=\varwidth, height = \varheight]{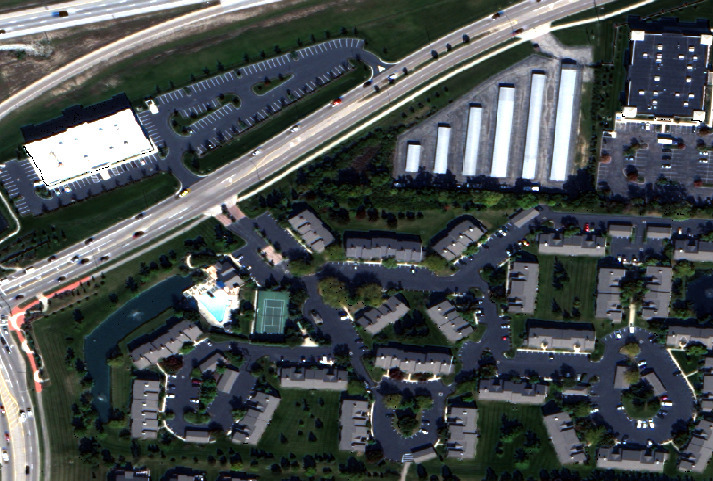}
  }
  \subfloat[Predicted Labels]
  {
    \includegraphics[width=\varwidth, height = \varheight]{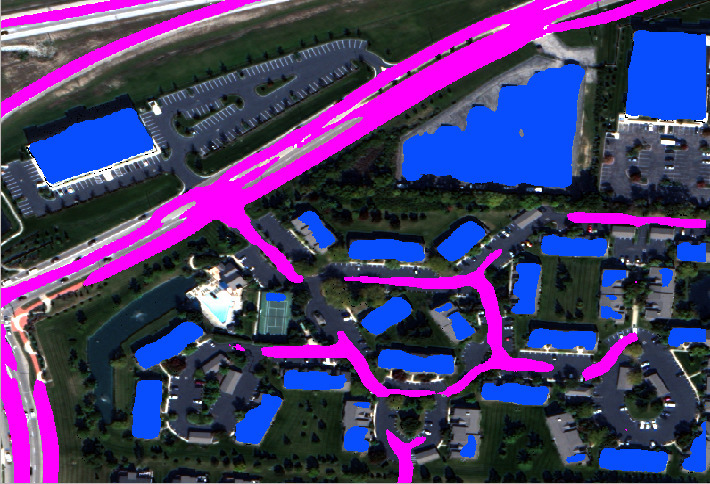}
  }\\
  \subfloat[Ortho View ]
  {
    \includegraphics[width=\varwidth, height = \varheight]{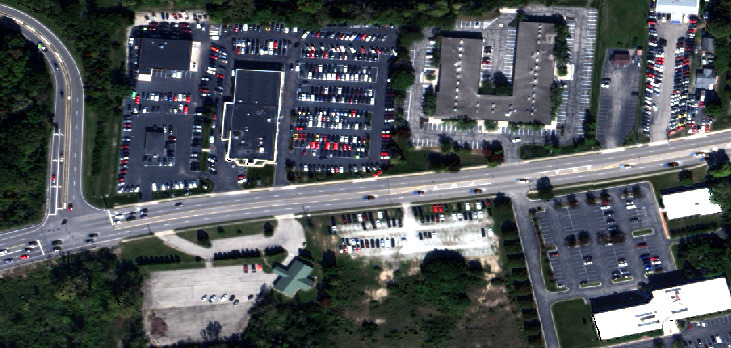}
  }
  \subfloat[Predicted Labels]
  {
    \includegraphics[width=\varwidth, height = \varheight]{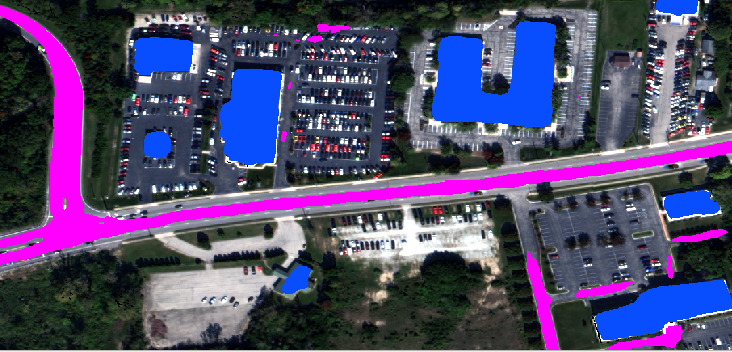}
  }\\
  \caption{Examples of orthorectified images and semantic
    labels output by our pipeline. Buildings are marked in
    \textcolor{blue}{blue} and roads are marked in
    \textcolor{magenta}{magenta}.}
  \label{fig:qual_eval_3}
\end{figure*}

\section{Conclusions}
\label{sec:conclusion}

We have presented a novel multi-view training paradigm that
significantly improves the accuracy of semantic labeling
over large geographic areas. The proposed approach
intelligently exploits information from multi-view and
multi-date images to provide robustness against noise in the
training labels. Our approach also speeds up inference, with
minimal increase in the GPU memory
requirements. Additionally, we have demonstrated that it is
possible to use OSM training data to reliably segment
large-area geographic regions and off-nadir satellite images
without any human supervision. While we have focused on
end-to-end automatic labeling of geographic areas, the ideas
put forth in this study can be incorporated into other
multi-view semantic-segmentation applications. Our research
opens up exciting possibilities for multi-view training in
related deep-learning tasks such as object detection and
panoptic segmentation.

With respect to semantic segmentation, one possible
  direction of future research is to design an architecture
  for multi-stage fusion of information from multiple
  views. More precisely, the features from different views
  could be combined at multiple layers of a CNN to yield
  possible improvements in accuracy. On a related note, one could also conduct a study to determine which layers of the SV CNN are influenced by the multi-view loss. Another exciting
  possibility would be to develop a multi-view framework for
  off-nadir images. This would require the use of lookup
  arrays to map between the pixels (in different images)
  that correspond to the same world point, and to correctly
  backpropagate gradients. Yet another research direction would be
  to create normalized DSMs and input them to the multi-view
  CNN framework. With respect to large-area image alignment
  and DSM creation, it might be advantageous to investigate
  the use of a second stage of alignment to resolve the generally small
  alignment differences between neighboring tiles. In
  addition, it should be possible to model the
  errors in the image-alignment and stereo-matching
  algorithms and subsequently use these models to
  construct more accurate DSMs. We plan to use our
  end-to-end automated framework to carry out these studies
  as part of future research.

\appendices
\section{Alignment of Full-Sized Satellite Images}

\subsection{Tiling}

The WorldView-3 images we have used in this study are
typically of size $43008 \times 38000$ in pixels and cover
an area of the ground of size 147 $km^2$.  In general,
images of this size must be broken into {\em image patches},
with each image patch covering a {\em tile} on the ground.
This is made necessary by the following three
considerations:

\begin{itemize}
\item As we describe in Appendix \ref{sec:image_align}, the
  corrections to the camera model calculated for
  high-precision alignment of the images with one another
  cannot be assumed to be constant across an entire
  satellite image.

\item
The image alignment algorithms usually start with the
extraction of {\em tie points} from the images. Tie points
are the corresponding key points (like those yielded by
interest operators like SIFT and SURF) in pairs of images.
The computational effort required for extracting the tie
points goes up quadratically as the size of the images
increases since the key points must be compared across
larger images.

\item
The run-time memory requirements of modern stereo matching
algorithms, such as those based on semi-global matching
(SGM), can become much too onerous for full sized satellite
images.

\end{itemize}

Based on our experience with WV3 images, we divide the
geographic area into overlapping tiles where each tile
consists of a central 1 $km^2$ main area and a 300 m overlap
with each of the four adjoining tiles.  This makes for a
total area of 2.56 $km^2$ for each tile.\footnote{A more
  accurate way to refer to a tile would be that it exists on
  a flat plane that is tangential to the WGS ellipsoid model
  of the earth.  This definition does not depend on whether
  the underlying terrain is flat or hilly.}  The image
patches that cover tiles of this size are typically of size
$5300 \times 5300$ in pixels.

Note that the notion of a tile is used only for aligning the
images and for constructing a DSM. This DSM is needed to
orthorectify the satellite images in order to bring them
into correspondence with OSM and with one another. For the
CNN-based machine-learning part of the system, we work
directly with the whole images and with the OSM for the
entire geographic area of interest.

\subsection {Image-to-Image Alignment}
\label{sec:image_align}

Aligning the satellite images that cover a geographic area
means that if we were to project a hypothetical ground point
into each of the images, the pixels thus obtained would
correspond to their actually recorded positions with
sub-pixel precision.  If this exercise were to be carried
out for WV3 images without first aligning them, the
projections in each of the satellite images could be off by
as much as 7 pixels from their true locations.

One needs a camera model for the images in order to
construct such projections and, for the case of WV3
images, the camera model comes in the form of rational
polynomial coefficients (RPCs).

\begin{figure*}
  \subfloat[]
  {
    \includegraphics[width=1\linewidth,height=0.4\linewidth]{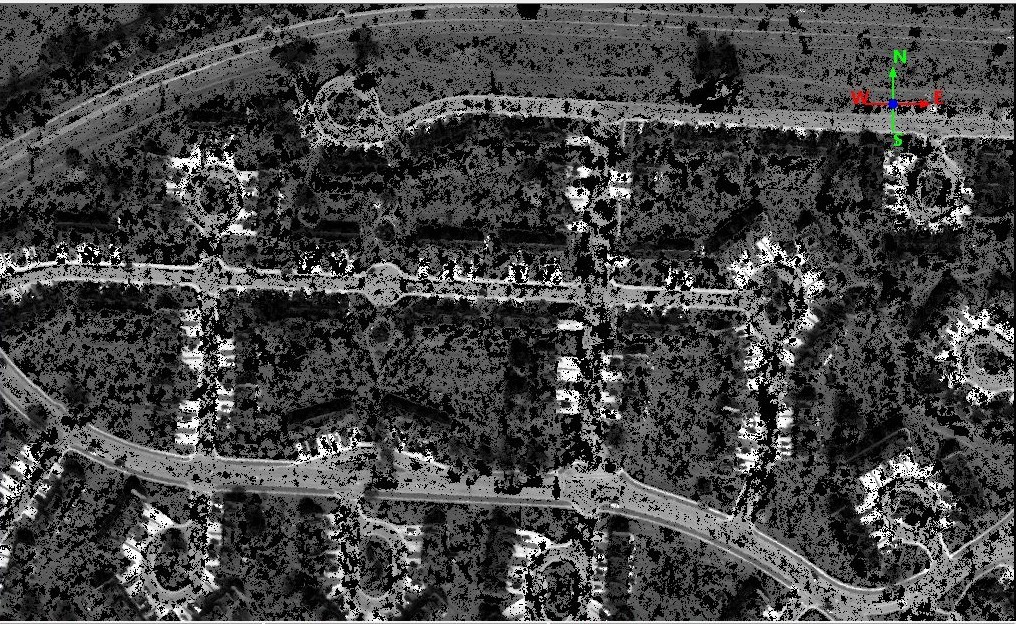}
  }\\
  \subfloat[]
  {
    \includegraphics[width=1\linewidth,height=0.4\linewidth]{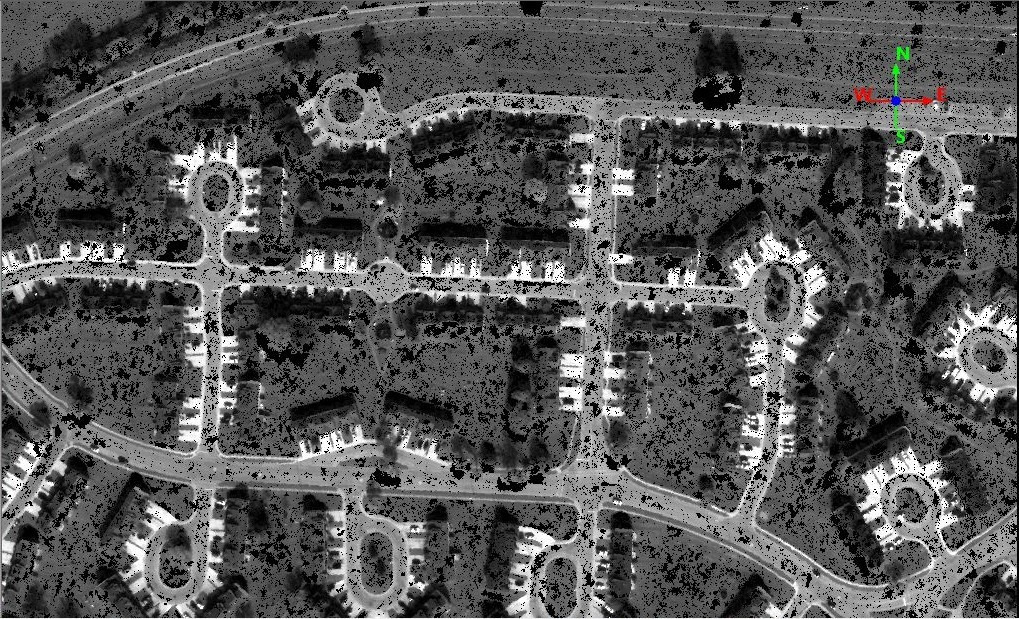}
  }
  \caption[An example to show why one cannot use a constant
  bias correction for a full-sized image]{An example to show
    why one cannot use a constant bias correction for a
    full-sized image. At top is the ortho view of a portion
    of a pairwise point cloud for the constant bias
    assumption. At bottom is the same for tile-based bias
    corrections. The points have been colored using the
    color from the images}
  \label{fig:var_bias}
\end{figure*}

It was shown by Grodecki and Dial \cite{Grodecki} that the
residual errors in the RPCs, on account of small
uncertainties in the measurements related to the position
and the attitude of a satellite, can be corrected by adding a
{\em constant bias} to the projected pixel coordinates of
the ground points, provided the area of interest on the
ground is not too large. We refer to this as the {\em
  constant bias assumption} for satellite image
alignment. We have tested the constant bias assumption
mentioned above and verified its validity for image patches
of size $5300 \times 5300$ (in pixels) for the WV3
images. Fig. \ref{fig:var_bias} presents evidence that the
constant bias assumption fails for a full-sized satellite
image.

\subsection{Tile-Based Alignment of Large-Area Satellite Images}
\label{sec:tiept_switch}

In order to operate on a large-area basis, we had to extend
the standard approach of bundle adjustment that is used to
align images.  The standard approach consists of: (1)
extracting the key points using an operator like SIFT/SURF;
(2) establishing correspondences between the key points in
pairs of images on the basis of the similarity of their
descriptor vectors; (3) using RANSAC to reject the outliers
in the set of correspondences (we refer to the surviving
correspondences as the {\em pairwise tie points}); and (4)
estimating the optimum bias corrections for each of the
images by the minimization of a reprojection-error based
cost function.

We have extended the standard approach by: (1) augmenting
the {\em pairwise tie points} with {\em multi-image tie
  points}; and (2) adding an $L_2$ regularizer to the
reprojection-error based cost function.  In what follows, we
start with the need for {\em multi-image tie points}.

Our experience has shown that doing bundle adjustment with
the usual pairwise tie points does not yield satisfactory
results when the sun angle is just above the horizon or when
there is significant snow-cover on the ground.  Under these
conditions, the decision thresholds one normally uses for
extracting the key points from the images often yield an
inadequate number of key points.  And if one were to lower the
decision thresholds, while that does increase the number of
key points, it also significantly increases the number of
false correspondences between them.

In such images, one gets better results overall by
extracting what we refer to as {\em multi-image tie points}.
The main idea in multi-image tie-point extraction is to
construct a graph of the key points detected with lower
decision thresholds and subsequently identify the key
points that correspond to the same putative world point
across multiple images, as opposed to just two
images.\footnote{The multi-image tie-point extraction module
  was developed by Dr. Tanmay Prakash.} Unfortunately,
multi-image tie points are computationally more expensive
than pairwise tie points --- roughly three times more
expensive.  Therefore, they must be used only when needed.

We have developed a ``detector'' that automatically
identifies the tiles that need the extra robustness provided
by the multi-image tie points.  The detector is based on the
rationale that the larger the extent to which each image
shares key-point correspondences with {\em all} the other images,
the more accurate the alignment is likely to be.  This
rationale is implemented by constructing an attributed graph
in which each vertex stands for an image and each edge for
the number of key-point correspondences between a pair of
images.  If we denote the largest connected component in this graph by
$C$, the extent to which each node in $C$ is connected with
all the other nodes in the same component can then be measured
by the following ``density'':
\begin{equation}
  D(C) = \frac{2|E_c|}{|C|(|C|-1)}
\end{equation} 

where $|E_c|$ is the total number of edges and $|C|$ is the
total number of vertices in $C$ respectively.  The detection
for the need for multi-image tie points is carried out by
first applying a threshold to $|C|$ and then to $D(C)$. This
detection algorithm is described in detail in
Fig. \ref{alg:image_align}. The algorithm is motivated by
the observation that a dense tie point graph based on
pairwise tie points is indicative of good alignment.

After the tie points --- pairwise or multi-image --- have
been identified in all the image patches for a given tile,
we apply sparse bundle adjustment (SBA) to them to align the
image patches.  The implementation of SBA includes an
$L_2$-regularization term that is added to the
reprojection-error based cost function because it
significantly increases the overall global accuracy of the
alignment. The only remaining issue with regard to the
alignment of the images is inter-tile alignment which we
discuss in Appendix \ref{sec:merge_dsm}.

\begin{figure*}
\begin{boxedalgorithmic}
  \STATE \hspace{0.2\columnwidth} \large{An Algorithm to Detect the Need for Multi-Image Tie Points}
  \par\noindent\rule[0.5\columnsep]{0.95\columnwidth}{0.01pt}
  \normalsize
  \linespread{1.2}\selectfont
  \STATE $S$ -- Total number of image patches to be aligned\\
  \textbf{Step 1: Run alignment using pairwise tie points}\\
  $T_p$ -- Tie-point graph returned by alignment using pairwise tie points\\
  $V$ -- Set of all image patches $\{v_i\}$. Each image patch is a vertex of $T_p$\\
  $E$ -- Set of all edges $\{e_{ij}\}$. $e_{ij}$ is an edge between the vertices $v_i$ and $v_j$ with a weight equal to the number of tie points between $v_i$ and $v_j$\\
  $k$, $D_{min}$ -- User-specified thresholds\\
  $AQ$ -- Flag set to True if alignment is of satisfactory quality. Otherwise set to False.\\
  \textbf{Evaluate alignment quality}
  \begin{enumerate}
  \item Find the largest connected component $C$ of $T_p$. \\$|C|$ is the number of image patches in $C$. $|E_c|$ is the number of edges in $C$.
  \item Check how many image patches have been aligned.\\ If $|C| < k \cdot S$, where $0 < k < 1$, $AQ \gets \textit{False}$. Return AQ
  \item Check if $C$ is a tree, i.e., if $|E_c| == |C| - 1$,
    $AQ \gets \textit{False}$. Return AQ \\ Explanation --
    The pushbroom camera model can be closely approximated
    by an affine camera model, i.e., the camera rays are
    almost parallel. Therefore, if $C$ is a tree, then for
    each pair of image patches, the two image patches might
    be well aligned with each other. However, distinct pairs
    might not be aligned with one another.
  \item Check the sparsity of $C$. $D(C)$ is the density of $C$. $D(C) =
    \frac{2|E_c|}{|C|(|C|-1)}$\\ 
    If $D(C) < D_{min}$, $AQ \gets \textit{False}$. Return AQ
  \end{enumerate}
  \textbf{Step 2: If $AQ == \textit{False}$, rerun alignment using multi-image tie points}
\end{boxedalgorithmic}
\caption{An algorithm to detect the need for multi-image tie
  points}
\label{alg:image_align}
\end{figure*}

\section{Creating a Tile-Level DSM}
\subsection{Stereo Matching}
\label{sec:pairwise_dsm}

As a first step towards constructing a DSM, stereo matching
is carried out in a pairwise manner. Similar to the study
reported in \cite{s2p}, pairs of images are selected based
on heuristics such as the difference in view angles,
difference in sun angles, time of acquisition, absolute view
angle, etc. In addition, images are selected to cover as
wide an azimuth-angle distribution as possible. We err on
the side of caution and select a minimum of 40 and a maximum
of 80 pairs per tile.  For each selected pair, the images
are rectified using the approach described by the study in
\cite{oh2010piecewise}.

For stereo matching, we use the hierarchical tSGM algorithm
\cite{rothermel2012sure} with some enhancements to improve
matching accuracy and speed. Specifically, we modify the
penalty parameters in the matching cost function as
described by the work in \cite{zbontar2016stereo}. We
noticed that this improves accuracy near the edges of
elevated structures. The second improvement is to use the
SRTM (Shuttle Radar Topography Mission) DEM (Digital
Elevation Model) \cite{srtm} that provides coarse terrain
elevation information at a low resolution (30 m). This DEM
does not contain the heights of buildings. We use the DEM to
better initialize the disparity search range for every point
in the disparity volume through a novel procedure that we
refer to as ``DEM-Sculpting''. Additional details regarding ``DEM-Sculpting'' can be found in our work described in \cite{demsculpt}.
This improves accuracy and speeds up stereo
matching. Additionally we use a guided bilateral filter for
post-processing. With these additions, the matching
algorithm is able to handle varying landscapes across a
large area.

\subsection{Pairwise Point-Cloud Creation and Fusion}
\label{sec:dsm_fusion}

The disparity maps and corrected RPCs are then used to
construct pairwise point clouds. Since the images have
already been aligned, the corresponding point clouds are
also aligned and can be fused without any further 3D
alignment. At each grid point in a tile, the median of the
top $Y$ values is retained as the height at that point,
where $Y$ is an empirically chosen parameter. Subsequently,
median filtering and morphological and boundary based
hole-filling techniques are applied.\footnote{The
  point-cloud generation and fusion modules as used in our
  framework were developed by John Papadakis from Applied
  Research Associates (ARA).}

\subsection{Merging Tile-Level DSMs}
\label{sec:merge_dsm}

On account of the high absolute alignment precision achieved
by using the $L_2$ regularization term in the bundle
adjustment logic, our experience shows that nothing further
needs to be done for merging the tile-level DSMs into a
larger DSM. To elaborate, the statistics of the differences
in the elevations at the tile boundaries are shown in Table
\ref{tbl:tile_edges} in Appendix \ref{sec:inter-tile}. We
see that the median absolute elevation difference at the
tile boundaries is less than 0.5 m -- an error that is much
too small to introduce noticeable errors in
orthorectification. 
We crop out the center 1 $km^2$ region from each DSM tile
and place it in the coordinate frame of the larger DSM. This
sidesteps the need to resolve any noise-induced variations
in the overlapping regions.

\section{Generating Training Data using Pansharpened Images
  and OSM}

\subsection{Pansharpening and Orthorectification}
Using the fused DSM as the elevation map, the system is now
ready for orthorectifying the satellite images
that cover the geographic area.  Orthorectification means
that you map the pixel values in the images to their
corresponding ground-based points in the geographic area of
interest.  What the system actually orthorectifies are the
pansharpened versions of the images --- these being the
highest resolution panchromatic images (meaning grayscale
images) that are assigned multispectral values from the
lower resolution multispectral data.

Orthorectification of an off-nadir image can lead to
``nodata'' regions on the ground if the pixels corresponding
to those regions are occluded in the image by tall
structures. Our system automatically delineates such regions
with a mask that is subsequently used during training of the
CNN to prevent gradients at those points from being
backpropagated. Each orthorectified image is resampled at a
resolution of 0.5 m. More details on orthorectification can
be found in Appendix \ref{sec:gwarp++}.

\subsection {Aligning OSM with Orthorectified Images}
\label{sec:osm_align_train_data_gen}
This module addresses the noise arising from any
misalignments between the OSM and the orthorectified
images. Our framework incorporates the following two
strategies to align the OSM with the orthorectified images:

\begin{enumerate}
\item Using Buildings: First, the system subtracts the DEM
  from the constructed DSM to extract coarse building
  footprints. Subsequently, these building footprints are
  used to align the orthorectified images with the OSM using
  Normalized Cross Correlation (NCC). This strategy has
  proved useful in areas with inadequate OSM road labels.

\item Using Roads: First, the system uses the ``Red Edge''
  and ``Coastal'' bands to calculate the Non-Homogeneous
  Feature Difference (NHFD) \cite{nhfd}, \cite{nhfd_2} for
  each point in the orthorectified image and subsequently
  applies thresholds to the NHFD values to detect coarse
  road footprints. The NHFD is calculated using the formula:
  \begin{equation}
\text{NHFD} = \frac{(\text{Red Edge -
      Coastal})}{(\text{Red Edge + Coastal})}
\end{equation}
  Subsequently, the roads (noisy obviously) are aligned with
  the OSM roads using NCC. The system uses this strategy in
  rural areas that may not contain the buildings needed for
  the previous approach to work.
\end{enumerate}

After alignment, the OSM vectors are converted to raster
format with the same resolution as in the orthorectified
images. Thus there is a label for each geographic point in
the orthorectified images. The OSM roads are thickened to
have a constant width of 8 m.

\begin{figure*}
  \centering{}
  \subfloat[]
  {\centering
    \includegraphics[width=0.4\linewidth,height=0.25\linewidth]{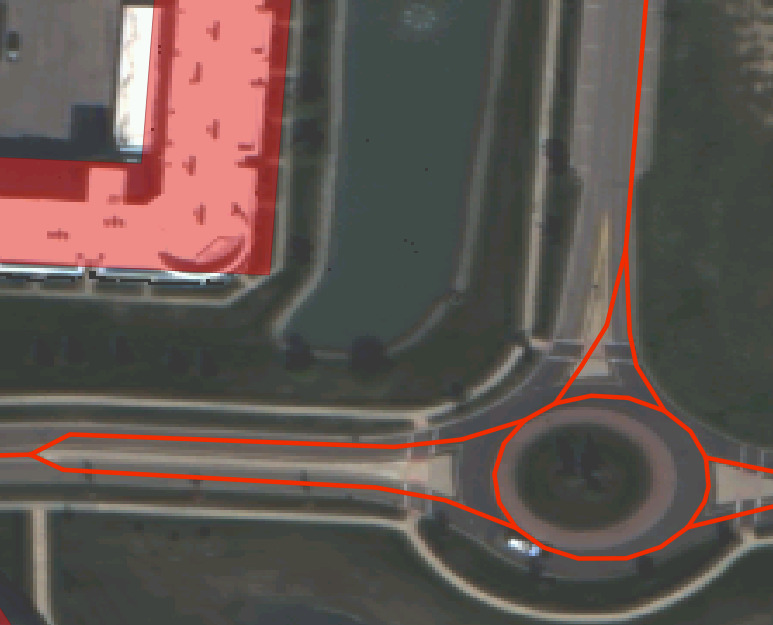}
  }
  \subfloat[]
  {
    \centering
    \includegraphics[width=0.4\linewidth,height=0.25\linewidth]{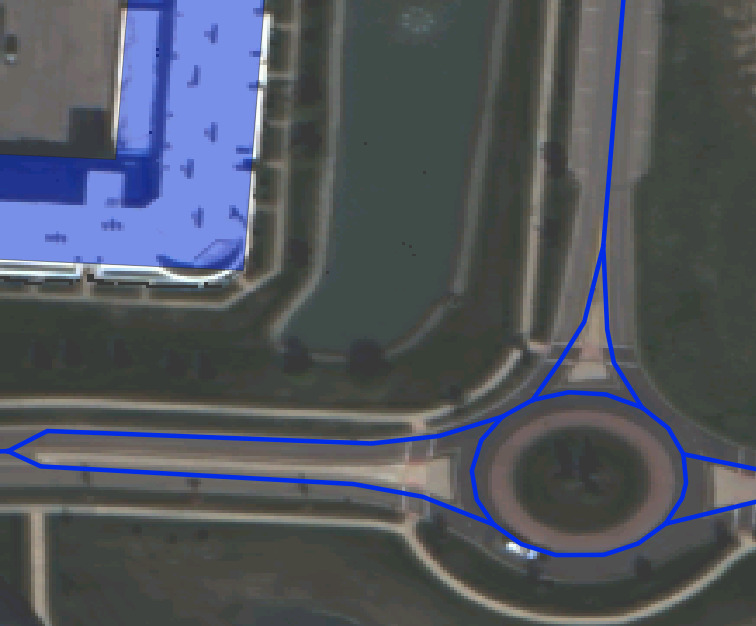}
  }
  \caption{This figure shows typical results obtained by
    aligning the orthorectified images with OSM.  What is
    shown in \textcolor{red}{red} at left are the unaligned
    OSM vectors, and what is in \textcolor{blue}{blue} at
    right are the aligned versions of the same.}
  \label{fig:osm_align}
\end{figure*}

Fig. \ref{fig:osm_align} shows misaligned and aligned OSM
vectors. It should be noted that some residual
alignment error does persist. We plan to improve this module
by aligning each building/road separately.

\section{True Orthorectification Using gwarp++}
\label{sec:gwarp++}

We can orthorectify the pansharpened images using the fused
DSM as the elevation map. Orthorectification is the process
of mapping the pixel values in the images to their
corresponding points in the geographic area of interest.
There is an important distinction to be made between
orthorectification and true orthorectification. If a LiDAR
point cloud or DSM is not available, the common practice is
to orthorectify images by using a DEM as the source of
elevation information. Since a DEM does not contain the
heights of elevated structures (buildings, trees, etc.),
such an orthorectified view will not represent a true nadir
view of the ground. For instance, the vertical walls of
buildings will be visible in such a view. To create a true
ortho view, we need to take the heights of the elevated
structures into account. While doing so, we need to detect
those portions of the scene that are occluded by taller
structures. Obviously these occluded portions will vary
depending upon the satellite view angle.

To the best of our knowledge, there are no open-source
utilities to create true ortho images using RPCs and DSMs at
this time. Therefore, we have developed a utility, which we
have named ``gwarp++'', to create full-sized true ortho
images quickly and efficiently. Interested researchers can
download the ``gwarp++'' software from the link at
\cite{gwarp}. We will now provide a brief overview of
``gwarp++''.

\begin{figure*}
\begin{boxedalgorithmic}
  \STATE \hspace{0.15\columnwidth} \large{An Algorithmic
    Description of ``gwarp++'' for True Orthorectification}
  \par\noindent\rule[0.5\columnsep]{0.95\columnwidth}{0.01pt}
  \normalsize
  \linespread{1.25}\selectfont
  \STATE $\phi$ -- Latitude, $\lambda$ -- Longitude, $h$ -- height\\
  $\phi_{\text{step}}$ -- Latitude step size,
  $\lambda_{\text{step}}$ -- Longitude step size,
  $h_{\text{step}}$ -- Height step size\\
  Image$_{\text{Patch}}$ -- Off-Nadir image patch\\
  $L$ -- Length of Image$_{\text{Patch}}$ in pixels, $W$ -- Width of Image$_{\text{Patch}}$ in pixels\\
  $LT \gets Zeros(W,L)$   \COMMENT{// Initialize lookup table to zeros}\\
  OutArray -- Output array for the orthorectified grid\\
  \textbf{Step 1: Find the extents of the tile spanned by the image patch}\\
  $(\phi_{\text{min}}, \lambda_{\text{max}})$ -- Top-left corner of the tile\\
  $(\phi_{\text{max}}, \lambda_{\text{min}})$ -- Bottom-right corner of the tile\\
  \textbf{Step 2: Project points into Image$_{\text{Patch}}$  and update $LT$}\\
  \FOR{$\phi = \phi_{\text{min}} \ ; \ \phi \ \leq \phi_{\text{max}} \ ; \ \phi = \phi + \phi_{\text{step}}$}
  {
    \FOR{$\lambda = \lambda_{\text{max}} \ ; \ \lambda \ \geq \lambda_{\text{min}} \ ; \ \lambda \ = \lambda - \lambda_{\text{step}}$}
    \STATE{
      $h \gets \text{DSM}(\phi, \lambda)$\\
      $h_{\text{ground}} \gets \text{DEM}(\phi, \lambda)$\\
      \FOR{$h' = \ h \ ; \ h' \geq h_{\text{ground}} \ ; \ h' = h' - h_{\text{step}}$}
      \STATE{
        $(s,l) \gets$ Proj$_{\text{RPC}}(\phi, \lambda, h')$ \COMMENT{// Proj$_{\text{RPC}}$ denotes the RPC equations used to project the 3D point into the image}\\ 
        \IF{$LT(s,l) < h'$} \STATE{ $LT(s,l) \gets h'$ }\ENDIF
      }\ENDFOR
    }\ENDFOR
  }\ENDFOR\\
  \textbf{Step 3: Create OutArray with a second pass over the grid}\\
  \FOR{$\phi = \phi_{\text{min}} \ ; \ \phi \ \leq
    \phi_{\text{max}} \ ; \ \phi = \phi +
    \phi_{\text{step}}$} {
    \FOR{$\lambda = \lambda_{\text{max}} \ ; \ \lambda \
      \geq \lambda_{\text{min}} \ ; \ \lambda \ = \lambda -
      \lambda_{\text{step}}$}\STATE{
      $h \gets \text{DSM}(\phi, \lambda)$\\
      $(s,l) \gets$ Proj$_{\text{RPC}}(\phi, \lambda, h)$\\
      \IF{$LT(s,l) > h + \gamma$}
      \STATE{OutArray$(\phi, \lambda) \gets $ NODATA}
      \ELSE
      \STATE{OutArray$(\phi, \lambda) \gets $ Image$_{\text{Patch}}(s,l)$ \COMMENT{// Can also interpolate values}
      }\ENDIF
    }\ENDFOR
  }\ENDFOR
\end{boxedalgorithmic}
\caption{An algorithmic description of ``gwarp++'' for true
  orthorectification}
\label{alg:gwarp++}
\end{figure*}

We will first discuss the case of orthorectifying an image
patch (that belongs to a single tile) with the help of a
DSM. Consider two points W$_1 = ( \phi_1, \lambda_1, h_1 )$
and W$_2 = ( \phi_2, \lambda_2, h_2 )$ that both project to
the same pixel coordinates in the image patch. $\phi$,
$\lambda$ and $h$ denote the latitude, longitude and height
coordinates respectively. If $h_2 > h_1$, it means that
W$_1$ is occluded by W$_2$. This is the core idea that
``gwarp++'' uses to detect the occluded ``nodata'' points.

Now, consider a single world point W $=(\phi, \lambda, h )$,
where $h$ is the height value from the DSM. Let
$h_{\text{ground}}$ be the corresponding height value in the
DEM. The DEM gives us a rough estimate of the height of the
ground. It is possible to use more sophisticated techniques,
such as the one described by the study in \cite{csf}, to
directly estimate the elevation of the ground from the
DSM. The DEM is sufficient for our application. Instead of
just projecting W into the image patch, ``gwarp++''
projects a set of points
\begin{align*}
& \hspace{0.25\linewidth} \mathcal{W'} = \big\{( \phi, \lambda, h')\big\} \
  & \\
  &\forall \ h' \ \epsilon \ [\ h, \ h - h_{\text{step}}, \ h -
  2 \cdot h_{\text{step}},...,\ h_{\text{ground}} \ ]&
\end{align*}
where $h_{\text{step}}$ is a user-defined step size.
$\mathcal{W'}$ is therefore a set of points sampled along
the vertical line from W to the ground. To understand the
motivation for doing this, it might help to consider the
case when W is the corner of the roof of a building. In that
case, $\mathcal{W'}$ is the set of points along the
corresponding vertical building edge from W to the
ground. If we apply this procedure to all the points on the
roof of a building, we will end up projecting the entire
building into the image patch.

We now describe the implementation of ``gwarp++'' below. The
algorithm is summarized in Fig. \ref{alg:gwarp++}.

\begin{enumerate}
 
\item ``gwarp++'' starts out by dividing the tile into a 2D
  grid of world points. The grid is 2D in the sense that
  only the longitude and the latitude coordinates are
  considered. The extents of this grid can be determined in
  an iterative fashion by using the RPC equations and the
  pixel coordinates of the corners of the image patch. The
  distance between the points of this grid is a user-defined
  parameter.

\item Using the height values from the DSM, for each point
  in the grid, ``gwarp++'' projects a set of points into
  the image patch as explained above. For each pixel in the
  image patch, a lookup table ``$LT$'' stores the maximum
  height, with the maximum being computed across all the
  points that project into this pixel. This procedure is
  repeated for all the points in the 2D grid.

\item At this stage, for each point J in the 2D grid, we
  know three things:
  \begin{itemize}
  \item $h_J$ -- The DSM height value at J
  \item $(s,l)$ -- The pixel into which J projects after
    assigning J an elevation value of $h_J$
  \item $LT(s,l)$ -- The maximum height of a world point
    that projects into $(s,l)$
  \end{itemize}
  If $LT(s,l) \ > h_J$, we can conclude that J is occluded by
  some other world point that has a height value of
  $LT(s,l)$.

\item Therefore, using a second pass over all the points of
  the 2D grid, ``gwarp++'' marks the occluded points
  with a ``NODATA'' value. In practice, to account for
  quantization errors and the noise in the DSM,
  ``gwarp++'' checks if $LT(s,l) \ > h_J \ + \ \gamma$
  where $\gamma$ is chosen appropriately.

\end{enumerate}

To orthorectify the full-sized image, we orthorectify each
image patch using its corrected RPCs and the large-area
DSM. The orthorectified image patches are then mosaiced into
a full-sized orthorectified image during which the
overlapping portions between the image patches are
discarded.

``gwarp++'' is written in C++. It has the nice property
of being massively parallel since the projection for each
point can be carried out independently and since each tile
can be processed independently. This parallelism is
exploited at both stages. For each image patch, OpenMP
\cite{openmp} is used to process the points in parallel. And
the different image patches are themselves orthorectified in
parallel by different virtual machines running on a
cloud-based framework.

For our application, each full-sized orthorectified image is
resampled at a resolution of 0.5 m. Furthermore, the occluded
points are delineated with a mask that is subsequently used
during training of the CNN to prevent gradients at those
points from being backpropagated.

\subsection{Accuracy of ``gwarp++''}

We conclude our discussion on true orthorectification with a
few remarks on the accuracy of the orthorectified images
produced by ``gwarp++''.

\textbf{3D vs 2.5D:} For each point W, ``gwarp++''
considers points along the vertical line from W to the
ground. This is not a good strategy for buildings that
possess more exotic shapes such as spherical water towers or
buildings with walls that slope inwards. In these cases
``gwarp++'' can incorrectly mark some points as occluded
points. The only way to handle such cases is by using a 3D
point cloud instead of a 2.5D DSM, which is beyond the scope
of our discussion.

\textbf{Error Propagation:} Errors in the RPCs and errors in
the DSM will translate into errors in the orthorectified
images. However, in our application, these errors are
largely drowned out by the errors in the OSM
labels. Nevertheless, it might be useful to study how these
errors propagate, which we leave for future work.

\section{Quantitative Evaluation of Alignment}
\label{sec:align_qual}
\subsection{Image-to-Image Alignment}
We use multiple metrics to evaluate the quality of
alignment. Table \ref{table:align_reproj} shows the average
reprojection error across tiles (and images) for both
regions, before and after alignment. Average reprojection
error goes down from 5-7 pixels to $\text{0.3}$ pixels for
both regions.

\begin{table}[H]
\setlength{\tabcolsep}{4pt}
\begin{center}
\caption{Average reprojection error in pixels across tiles and images in Ohio and California}
\label{table:align_reproj}
\renewcommand{\arraystretch}{1.5}
\begin{tabular}{|c|c|c|c|}
  \hline
  Region & & Mean & Variance \\
  \hline\hline
  \multirow{2}{4em}{Ohio} 
         & Unaligned & 6.70 & 0.180 \\
  \cline{2-4}
         & Aligned & {\bf 0.30} & 0.003 \\
  \hline
  \multirow{2}{4em}{California} 
         & Unaligned & 5.71 & 0.280 \\
  \cline{2-4} 
         & Aligned & {\bf 0.32} & 0.001 \\
  \hline
\end{tabular}
\end{center}
\end{table}

Since pushbroom sensors can be closely approximated by
affine cameras with parallel rays, reprojection error alone
does not give the complete picture. For our second metric,
we manually annotate tie points in 31 out of 32 images over
a 1 km$^2$ region in Ohio and in all 32 images over a 2
km$^2$ region in California.  Within these regions, we
measure the pairwise alignment errors for all possible pairs
of images and report them in Table
\ref{table:align_pairwise}. One can observe that most of the
pairs are aligned with subpixel error. This is a much harder
metric than the mean reprojection error. It is important to
use this metric especially since stereo matching requires
subpixel alignment accuracy. The good quality of alignment
across the large region is also reflected in the high
quality of the DSM and the semantic labeling metrics.

\begin{table}[H]
\begin{center}
\caption{Pairwise alignment error statistics using manually annotated groundtruth for Ohio and California}
\label{table:align_pairwise}
\renewcommand{\arraystretch}{1.5}
\begin{tabular}{|c|p{2cm}|p{2cm}|p{1.5cm}|}
  \hline
  Region & No. of pairs with error $<1$ pixel & No. of pairs with error $<2$ pixels & Total No. of pairs \\
  \hline\hline
  Ohio & 417 & 455 & 465 \\
  \hline
  California & 484 & 496 & 496 \\
  \hline
\end{tabular}
\end{center}
\end{table}

\subsection{Inter-Tile Alignment}
\label{sec:inter-tile}

Due to the high absolute alignment precision achieved by
using the $L_2$ regularization term in the bundle-adjustment
logic, it turns out that the tile-level DSMs are
well aligned with one another. Table \ref{tbl:tile_edges}
shows the statistics of the differences in the elevations at
the tile boundaries. It can be seen that the median absolute
elevation difference at the tile boundaries is less than 0.5
m -- an error that is much too small to introduce noticeable
errors in orthorectification.

\begin{table}[H]
\caption{Median of the absolute differences in elevation, and median of the
  rms value of the differences in elevation at the tile boundaries}
\label{tbl:tile_edges}
\renewcommand{\arraystretch}{1.5}
\begin{center}
\begin{tabular}{|c|p{2.8cm}|p{3cm}|}
  \hline
  Region & Median absolute Z diff & Median RMS of Z diff \\
  \hline\hline
  Ohio & 0.42 m & 0.72 m \\
  \hline
  California & 0.47 m & 0.79 m \\
  \hline
\end{tabular}
\end{center}
\end{table}

\section{A Distributed Workflow for Stereo Matching and DSM
  Creation}
\label{sec:dist_stereo_matching}

Creating DSMs for a 100 km$^2$ region is the most
computationally-intensive and the slowest module in the
framework shown in Fig.~\ref{fig:overview}. It is also the
module that is most likely to cause ``out-of-memory''
errors. Therefore, we need to carefully choose some specific
design attributes for this module, which we will highlight
in this section.

\begin{figure*}[htbp!]
  \centering
  \includegraphics[width=1\linewidth,height=0.5\linewidth]{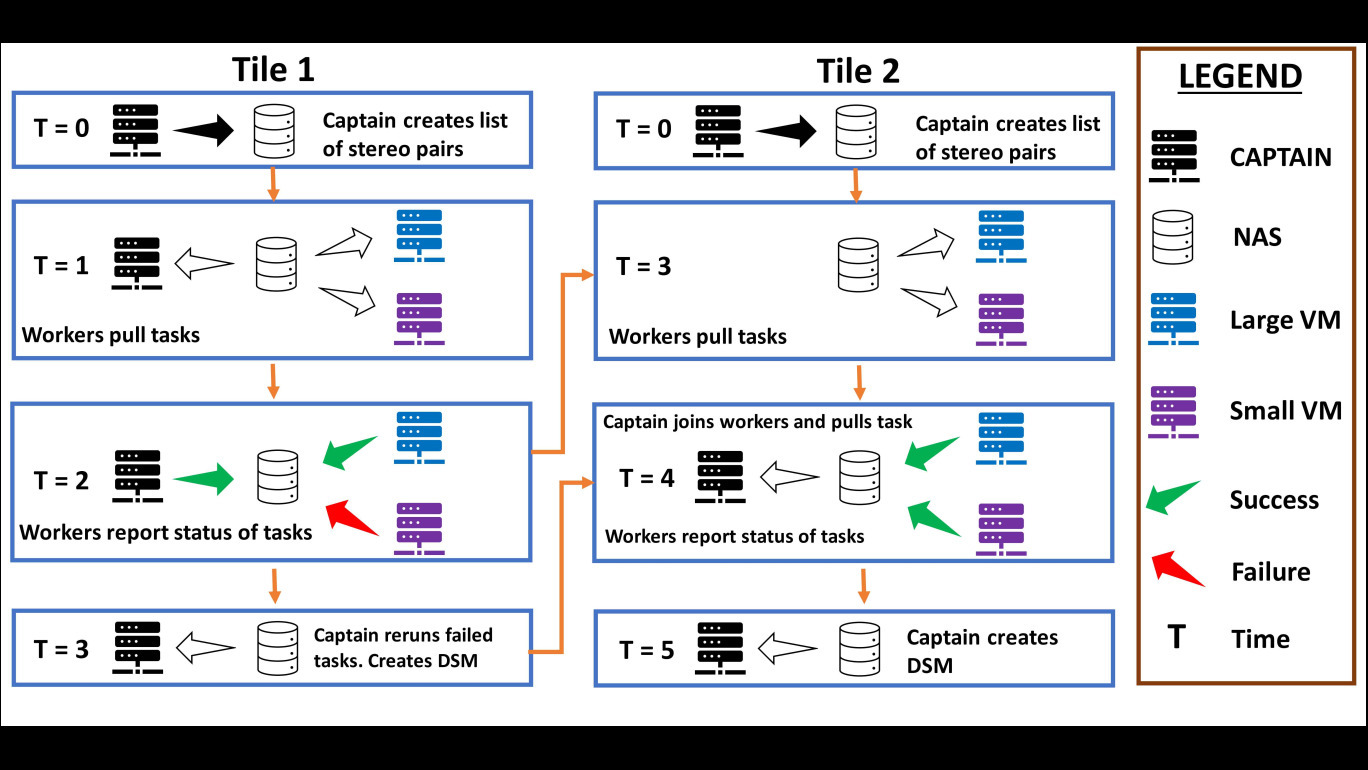}
  \caption{An example to illustrate our distributed
    stereo-matching and DSM-creation workflow. In this
    example, there are only 2 tiles and 3 selected stereo
    pairs for each tile. There are only 3 VMs, a captain, a
    small VM and a large VM.  T indicates the time
    stamp. Notice how at T = 3, two of the VMs have moved
    onto Tile 2 whereas the captain stays back to finish
    processing Tile 1.}
  \label{fig:distributed_stereo}
\end{figure*}

We can leverage the inherent parallelism in stereo matching
and in DSM creation by intelligently distributing the tasks
across a cloud computing system. The steps for distributed
stereo matching and DSM creation are enumerated below and
shown in Fig.~\ref{fig:distributed_stereo}.

\begin{enumerate}

\item A captain virtual machine (VM) prepares a list of the
  selected stereo pairs of image patches for each tile. This
  is done for all the tiles at the beginning. All the tiles
  are added to a queue. All the lists are stored on a shared
  Network Attached Storage (NAS).

\item The captain sends a message to all the worker VMs to
  start. The captain also assumes the role of a worker at
  this step.

\item \label{item:step2} For the first tile in the queue,
  the workers pull/request a pair of image patches to
  process. Safeguards are imposed to ensure that each worker
  gets a unique pair.
  
\item Each worker attempts to create a pairwise point cloud
  and subsequently reports the status of its task. Each
  worker then pull/requests the next unprocessed stereo pair
  for the current tile. Successfully processed stereo pairs
  are marked as done.

\item If there are no more unprocessed stereo pairs for this
  tile then:
  \begin{enumerate}[label=(\roman*)]
  \item The current tile is removed from the queue. All the
    idle workers except for the captain and the large VMs
    move on to the next tile in the queue, i.e., to step
    \ref{item:step2}. By a large VM, we mean a VM with more
    memory and a larger number of CPUs.
  \item All the stereo pairs for which point-cloud creation
    failed are processed for a second time by the remaining
    workers. Even if processing fails again, these stereo
    pairs are marked as done.
  \end{enumerate}

\item At this stage, all the selected stereo pairs for the
  current tile are marked as done. The large VMs join their
  smaller counterparts on the next tile, i.e., at step
  \ref{item:step2}. The captain alone starts the process of
  fusing the multiple pairwise point clouds into a single
  fused DSM for the current tile. After this, the captain
  also proceeds to join the other VMs in step
  \ref{item:step2}.
  
\end{enumerate}

A graphic illustration of the above workflow is shown in
Fig.~\ref{fig:distributed_stereo}. For the sake of clarity,
in this illustration, we assume that there are only 2 tiles
and that there are only 3 selected stereo pairs for each
tile. We also assume that there are only 3 VMs, a captain, a
small VM and a large VM.

\subsection{Advantages of this Distributed Workflow}

\begin{itemize}
\item No VM remains idle except during the last processing
  stage of the very last tile.
\item Failed pairs are processed twice to handle
  ``out-of-memory'' errors.
\item The intensive process of creating a fused DSM is
  carried out on the most powerful captain VM.
\item Note that we could have opted to use a simpler
  workflow where all the VMs wait for a fused DSM to be
  created before proceeding to the next tile. However, our
  workflow reduces the processing time by a number of
  days.

  For an example, assume that there are 10 VMs and 100
  tiles. Also assume that each stereo pair takes 20 minutes
  to process, that we process 80 pairs per tile and that the
  point-cloud fusion takes 60 minutes. If all the workers
  waited for a tile-level DSM to be created before moving
  on to the next tile, then it would take
  $ \frac{(\frac{80 \times 20}{10} + 60)}{60}
  \approx $ 3 hours and 40 minutes to finish processing a
  single tile. For 100 tiles it would take $\approx$ 15 days
  and 6 hours.

  Our workflow takes
  $ \frac{(\frac{80 \times 20}{10})}{60}
  \approx $ 2 hours and 40 minutes for a single tile. This
  is because while the captain is fusing the point clouds
  for a tile, the other VMs will be processing the next
  tile. For 100 tiles it would take $\approx$ 11 days and 2
  hours, roughly saving us 4 days of processing time.
\end{itemize}




\ifCLASSOPTIONcaptionsoff
  \newpage
\fi



\bibliographystyle{IEEEtran}
\bibliography{myieee}

\providecommand*\hyphen{-}
\begin{thebibliography}{10}
\providecommand{\url}[1]{#1}
\csname url@samestyle\endcsname
\providecommand{\newblock}{\relax}
\providecommand{\bibinfo}[2]{#2}
\providecommand{\BIBentrySTDinterwordspacing}{\spaceskip=0pt\relax}
\providecommand{\BIBentryALTinterwordstretchfactor}{4}
\providecommand{\BIBentryALTinterwordspacing}{\spaceskip=\fontdimen2\font plus
\BIBentryALTinterwordstretchfactor\fontdimen3\font minus
  \fontdimen4\font\relax}
\providecommand{\BIBforeignlanguage}[2]{{%
\expandafter\ifx\csname l@#1\endcsname\relax
\typeout{** WARNING: IEEEtran.bst: No hyphenation pattern has been}%
\typeout{** loaded for the language `#1'. Using the pattern for}%
\typeout{** the default language instead.}%
\else
\language=\csname l@#1\endcsname
\fi
#2}}
\providecommand{\BIBdecl}{\relax}
\BIBdecl

\bibitem{osm}
{OpenStreetMap contributors}, ``{Planet dump retrieved from
  \url{https://planet.osm.org} },'' \url{https://www.openstreetmap.org }, 2017.

\bibitem{gwarp}
``{gwarp++},'' \url{https://github.com/shallowlearn/gwarp}, accessed:
  2020-11-25.

\bibitem{flyby}
``{Flyby Videos of Large-Area DSMs},''
  \url{https://engineering.purdue.edu/RVL/CORE3D/DSM/}, accessed: 2020-11-25.

\bibitem{WV3}
``{WorldView\hyphen3},''
  \url{https://www.satimagingcorp.com/satellite-sensors/worldview-3/},
  accessed: 2018-08-06.

\bibitem{polymapper}
Z.~Li, J.~D. Wegner, and A.~Lucchi, ``{Topological map extraction from overhead
  images},'' in \emph{{Proceedings of the IEEE International Conference on
  Computer Vision}}, 2019, pp. 1715--1724.

\bibitem{roadtracer}
F.~Bastani, S.~He, S.~Abbar, M.~Alizadeh, H.~Balakrishnan, S.~Chawla,
  S.~Madden, and D.~DeWitt, ``{Roadtracer: Automatic extraction of road
  networks from aerial images},'' in \emph{{Proceedings of the IEEE Conference
  on Computer Vision and Pattern Recognition}}, 2018, pp. 4720--4728.

\bibitem{bastani2018machine}
F.~Bastani, S.~He, S.~Abbar, M.~Alizadeh, H.~Balakrishnan, S.~Chawla, and
  S.~Madden, ``{Machine-assisted map editing},'' in \emph{{Proceedings of the
  26th ACM SIGSPATIAL International Conference on Advances in Geographic
  Information Systems}}, 2018, pp. 23--32.

\bibitem{Chu_2019_ICCV}
H.~Chu, D.~Li, D.~Acuna, A.~Kar, M.~Shugrina, X.~Wei, M.-Y. Liu, A.~Torralba,
  and S.~Fidler, ``{Neural Turtle Graphics for Modeling City Road Layouts},''
  in \emph{{The IEEE International Conference on Computer Vision (ICCV)}},
  October 2019.

\bibitem{yang2019road}
X.~Yang, X.~Li, Y.~Ye, R.~Y. Lau, X.~Zhang, and X.~Huang, ``{Road Detection and
  Centerline Extraction Via Deep Recurrent Convolutional Neural Network
  U-Net},'' \emph{IEEE Transactions on Geoscience and Remote Sensing}, vol.~57,
  no.~9, pp. 7209--7220, 2019.

\bibitem{park2019refining}
E.~Park \emph{et~al.}, ``{Refining inferred road maps using GANs},'' Ph.D.
  dissertation, Massachusetts Institute of Technology, 2019.

\bibitem{Etten_2020_WACV}
A.~V. Etten, ``{City-Scale Road Extraction from Satellite Imagery v2: Road
  Speeds and Travel Times},'' in \emph{{The IEEE Winter Conference on
  Applications of Computer Vision (WACV)}}, March 2020.

\bibitem{mosinska2019joint}
A.~Mosinska, M.~Kozinski, and P.~Fua, ``{Joint Segmentation and Path
  Classification of Curvilinear Structures},'' \emph{IEEE transactions on
  pattern analysis and machine intelligence}, 2019.

\bibitem{yang2019road_v2}
X.~Yang, X.~Li, Y.~Ye, X.~Zhang, H.~Zhang, X.~Huang, and B.~Zhang, ``{Road
  Detection via Deep Residual Dense U-Net},'' in \emph{2019 International Joint
  Conference on Neural Networks (IJCNN)}.\hskip 1em plus 0.5em minus
  0.4em\relax IEEE, 2019, pp. 1--7.

\bibitem{eth_cnn}
P.~Kaiser, J.~D. Wegner, A.~Lucchi, M.~Jaggi, T.~Hofmann, and K.~Schindler,
  ``{Learning Aerial Image Segmentation From Online Maps},'' \emph{IEEE
  Transactions on Geoscience and Remote Sensing}, vol.~55, no.~11, pp.
  6054--6068, Nov 2017.

\bibitem{resunet}
Z.~Zhang, Q.~Liu, and Y.~Wang, ``{Road Extraction by Deep Residual U-Net},''
  \emph{IEEE Geoscience and Remote Sensing Letters}, vol.~15, no.~5, pp.
  749--753, May 2018.

\bibitem{saito}
S.~Saito, Y.~Yamashita, and Y.~Aoki, ``{Multiple Object Extraction from Aerial
  Imagery with Convolutional Neural Networks},'' vol.~60, pp.
  10\,402--1/10\,402, 01 2016.

\bibitem{forez}
E.~Maggiori, Y.~Tarabalka, G.~Charpiat, and P.~Alliez, ``{Convolutional Neural
  Networks for Large-Scale Remote-Sensing Image Classification},'' \emph{IEEE
  Transactions on Geoscience and Remote Sensing}, vol.~55, no.~2, pp. 645--657,
  Feb 2017.

\bibitem{mnih_hybrid}
Y.~{Li}, L.~{Guo}, J.~{Rao}, L.~{Xu}, and S.~{Jin}, ``{Road Segmentation Based
  on Hybrid Convolutional Network for High-Resolution Visible Remote Sensing
  Image},'' \emph{IEEE Geoscience and Remote Sensing Letters}, vol.~16, no.~4,
  pp. 613--617, April 2019.

\bibitem{DeepVGI}
J.~Chen and A.~Zipf, ``{DeepVGI: Deep Learning with Volunteered Geographic
  Information},'' in \emph{WWW}, 2017.

\bibitem{OSMDeepOD}
\BIBentryALTinterwordspacing
S.~Kurath, ``\BIBforeignlanguage{de}{{OSMDeepOD \hyphen Object Detection on
  Orthophotos with and for VGI}},'' vol. Volume 2, pp. 173--188, online
  available: http://www.austriaca.at/?arp=0x00373589 - Last access:12.8.2018.
  [Online]. Available: \url{http://www.austriaca.at/?arp=0x00373589}
\BIBentrySTDinterwordspacing

\bibitem{DeepOSM}
``{DeepOSM},'' \url{https://github.com/trailbehind/DeepOSM}, accessed:
  2018-08-06.

\bibitem{gans}
\BIBentryALTinterwordspacing
I.~Goodfellow, J.~Pouget-Abadie, M.~Mirza, B.~Xu, D.~Warde-Farley, S.~Ozair,
  A.~Courville, and Y.~Bengio, ``{Generative Adversarial Nets},'' in
  \emph{Advances in Neural Information Processing Systems}, Z.~Ghahramani,
  M.~Welling, C.~Cortes, N.~Lawrence, and K.~Q. Weinberger, Eds.,
  vol.~27.\hskip 1em plus 0.5em minus 0.4em\relax Curran Associates, Inc.,
  2014, pp. 2672--2680. [Online]. Available:
  \url{https://proceedings.neurips.cc/paper/2014/file/5ca3e9b122f61f8f06494c97b1afccf3-Paper.pdf}
\BIBentrySTDinterwordspacing

\bibitem{deeproadmapper}
G.~{M\'attyus}, W.~{Luo}, and R.~{Urtasun}, ``{DeepRoadMapper: Extracting Road
  Topology from Aerial Images},'' in \emph{2017 IEEE International Conference
  on Computer Vision (ICCV)}, Oct 2017, pp. 3458--3466.

\bibitem{dlinknet}
L.~Zhou, C.~Zhang, and M.~Wu, ``{D-LinkNet: LinkNet With Pretrained Encoder and
  Dilated Convolution for High Resolution Satellite Imagery Road Extraction},''
  in \emph{CVPR Workshops}, 2018, pp. 182--186.

\bibitem{batra2019improved}
A.~Batra, S.~Singh, G.~Pang, S.~Basu, C.~Jawahar, and M.~Paluri, ``{Improved
  road connectivity by joint learning of orientation and segmentation},'' in
  \emph{Proceedings of the IEEE Conference on Computer Vision and Pattern
  Recognition}, 2019, pp. 10\,385--10\,393.

\bibitem{singh2018self}
S.~Singh, A.~Batra, G.~Pang, L.~Torresani, S.~Basu, M.~Paluri, and C.~Jawahar,
  ``{Self-Supervised Feature Learning for Semantic Segmentation of Overhead
  Imagery},'' in \emph{BMVC}, vol.~1, no.~2, 2018, p.~4.

\bibitem{rotich2018using}
G.~Rotich, S.~Aakur, R.~Minetto, M.~P. Segundo, and S.~Sarkar, ``{Using
  Semantic Relationships among Objects for Geospatial Land Use
  Classification},'' in \emph{2018 IEEE Applied Imagery Pattern Recognition
  Workshop (AIPR)}.\hskip 1em plus 0.5em minus 0.4em\relax IEEE, 2018, pp.
  1--7.

\bibitem{rotich2018resource}
G.~Rotich, R.~Minetto, and S.~Sarkar, ``{Resource-constrained simultaneous
  detection and labeling of objects in high-resolution satellite images},''
  \emph{arXiv preprint arXiv:1810.10110}, 2018.

\bibitem{costea2018roadmap}
D.~Costea, A.~Marcu, E.~Slusanschi, and M.~Leordeanu, ``{Roadmap Generation
  Using a Multi-Stage Ensemble of Deep Neural Networks With Smoothing-Based
  Optimization.}'' in \emph{CVPR Workshops}, 2018, pp. 220--224.

\bibitem{deepglobe}
\BIBentryALTinterwordspacing
I.~Demir, K.~Koperski, D.~Lindenbaum, G.~Pang, J.~Huang, S.~Basu, F.~Hughes,
  D.~Tuia, and R.~Raskar, ``{DeepGlobe 2018: {A} Challenge to Parse the Earth
  through Satellite Images},'' \emph{CoRR}, vol. abs/1805.06561, 2018.
  [Online]. Available: \url{http://arxiv.org/abs/1805.06561}
\BIBentrySTDinterwordspacing

\bibitem{spacenet}
``{SpaceNet on Amazon Web Services (AWS). "Datasets." The SpaceNet Catalog.
  Last modified April 30, 2018.}''
  \url{https://spacenetchallenge.github.io/datasets/datasetHomePage.html},
  accessed: 2018-08-06.

\bibitem{jhu_us3d}
M.~Bosch, K.~Foster, G.~Christie, S.~Wang, G.~D. Hager, and M.~Brown,
  ``{Semantic stereo for incidental satellite images},'' in \emph{2019 IEEE
  Winter Conference on Applications of Computer Vision (WACV)}.\hskip 1em plus
  0.5em minus 0.4em\relax IEEE, 2019, pp. 1524--1532.

\bibitem{jhu_urban3d}
\BIBentryALTinterwordspacing
H.~R. Goldberg, S.~Wang, G.~A. Christie, and M.~Z. Brown, ``{Urban 3D
  challenge: building footprint detection using orthorectified imagery and
  digital surface models from commercial satellites},'' in \emph{{Geospatial
  Informatics, Motion Imagery, and Network Analytics VIII}}, K.~Palaniappan,
  P.~J. Doucette, and G.~Seetharaman, Eds., vol. 10645, International Society
  for Optics and Photonics.\hskip 1em plus 0.5em minus 0.4em\relax SPIE, 2018,
  pp. 12 -- 31. [Online]. Available: \url{https://doi.org/10.1117/12.2304682}
\BIBentrySTDinterwordspacing

\bibitem{jhu_new3d}
\BIBentryALTinterwordspacing
M.~Brown, H.~Goldberg, K.~Foster, A.~Leichtman, S.~Wang, S.~Hagstrom, M.~Bosch,
  and S.~Almes, ``{Large-scale public lidar and satellite image data set for
  urban semantic labeling},'' in \emph{{Laser Radar Technology and Applications
  XXIII}}, M.~D. Turner and G.~W. Kamerman, Eds., vol. 10636, International
  Society for Optics and Photonics.\hskip 1em plus 0.5em minus 0.4em\relax
  SPIE, 2018, pp. 154 -- 167. [Online]. Available:
  \url{https://doi.org/10.1117/12.2304403}
\BIBentrySTDinterwordspacing

\bibitem{datafusion}
``{IEEE Data Fusion Contest},''
  \url{http://www.grss-ieee.org/community/technical-committees/data-fusion/data-fusion-contest/},
  accessed: 2018-08-06.

\bibitem{datafusioncon1}
A.~Lagrange, B.~L. Saux, A.~Beaup\`ere, A.~Boulch, A.~Chan-Hon-Tong, S.~Herbin,
  H.~Randrianarivo, and M.~Ferecatu, ``{Benchmarking classification of
  earth-observation data: From learning explicit features to convolutional
  networks},'' in \emph{2015 IEEE International Geoscience and Remote Sensing
  Symposium (IGARSS)}, July 2015, pp. 4173--4176.

\bibitem{isprs2d}
``{ISPRS-2D Dataset},''
  \url{http://www2.isprs.org/commissions/comm3/wg4/semantic-labeling.html},
  accessed: 2018-08-06.

\bibitem{torontocity}
\BIBentryALTinterwordspacing
S.~Wang, M.~Bai, G.~M{\'{a}}ttyus, H.~Chu, W.~Luo, B.~Yang, J.~Liang,
  J.~Cheverie, S.~Fidler, and R.~Urtasun, ``{TorontoCity: Seeing the World with
  a Million Eyes},'' \emph{CoRR}, vol. abs/1612.00423, 2016. [Online].
  Available: \url{http://arxiv.org/abs/1612.00423}
\BIBentrySTDinterwordspacing

\bibitem{dota}
\BIBentryALTinterwordspacing
G.~Xia, X.~Bai, J.~Ding, Z.~Zhu, S.~J. Belongie, J.~Luo, M.~Datcu, M.~Pelillo,
  and L.~Zhang, ``{{DOTA:} {A} Large-scale Dataset for Object Detection in
  Aerial Images},'' \emph{CoRR}, vol. abs/1711.10398, 2017. [Online].
  Available: \url{http://arxiv.org/abs/1711.10398}
\BIBentrySTDinterwordspacing

\bibitem{GRID}
``{Geospatial Repository and Data Management System},''
  \url{https://griduc.rsgis.erdc.dren.mil/griduc}, accessed: 2018-08-06.

\bibitem{datafusioncon2}
N.~{Yokoya}, P.~{Ghamisi}, J.~{Xia}, S.~{Sukhanov}, R.~{Heremans},
  I.~{Tankoyeu}, B.~{Bechtel}, B.~{Le Saux}, G.~{Moser}, and D.~{Tuia}, ``{Open
  Data for Global Multimodal Land Use Classification: Outcome of the 2017 IEEE
  GRSS Data Fusion Contest},'' \emph{IEEE Journal of Selected Topics in Applied
  Earth Observations and Remote Sensing}, vol.~11, no.~5, pp. 1363--1377, May
  2018.

\bibitem{mv-paper1}
L.~{Ma}, J.~{St\"uckler}, C.~{Kerl}, and D.~{Cremers}, ``{Multi-view deep
  learning for consistent semantic mapping with RGB-D cameras},'' in
  \emph{{2017 IEEE/RSJ International Conference on Intelligent Robots and
  Systems (IROS)}}, Sep. 2017, pp. 598--605.

\bibitem{mv-paper2}
J.~{Xiao} and L.~{Quan}, ``{Multiple view semantic segmentation for street view
  images},'' in \emph{{2009 IEEE 12th International Conference on Computer
  Vision}}, Sep. 2009, pp. 686--693.

\bibitem{mv-paper3}
L.~Ge, H.~Liang, J.~Yuan, and D.~Thalmann, ``{Robust 3D Hand Pose Estimation in
  Single Depth Images: From Single-View CNN to Multi-View CNNs},'' in
  \emph{{The IEEE Conference on Computer Vision and Pattern Recognition
  (CVPR)}}, June 2016.

\bibitem{mv-paper4}
H.~Su, S.~Maji, E.~Kalogerakis, and E.~Learned-Miller, ``{Multi-View
  Convolutional Neural Networks for 3D Shape Recognition},'' in \emph{{The IEEE
  International Conference on Computer Vision (ICCV)}}, December 2015.

\bibitem{mv-paper5}
C.~R. Qi, H.~Su, M.~Niessner, A.~Dai, M.~Yan, and L.~J. Guibas, ``{Volumetric
  and Multi-View CNNs for Object Classification on 3D Data},'' in \emph{{The
  IEEE Conference on Computer Vision and Pattern Recognition (CVPR)}}, June
  2016.

\bibitem{mv-paper6}
A.~Mortazi, R.~Karim, K.~Rhode, J.~Burt, and U.~Bagci, ``{CardiacNET:
  Segmentation of Left Atrium and Proximal Pulmonary Veins from MRI Using
  Multi-view CNN},'' in \emph{{Medical Image Computing and Computer-Assisted
  Intervention, MICCAI 2017}}, M.~Descoteaux, L.~Maier-Hein, A.~Franz,
  P.~Jannin, D.~L. Collins, and S.~Duchesne, Eds.\hskip 1em plus 0.5em minus
  0.4em\relax Cham: Springer International Publishing, 2017, pp. 377--385.

\bibitem{mv-paper8}
G.~Carneiro, J.~Nascimento, and A.~P. Bradley, ``{Unregistered Multiview
  Mammogram Analysis with Pre-trained Deep Learning Models},'' in
  \emph{{Medical Image Computing and Computer-Assisted Intervention -- MICCAI
  2015}}, N.~Navab, J.~Hornegger, W.~M. Wells, and A.~F. Frangi, Eds.\hskip 1em
  plus 0.5em minus 0.4em\relax Cham: Springer International Publishing, 2015,
  pp. 652--660.

\bibitem{mv-paper9}
J.~Han, H.~Chen, N.~Liu, C.~Yan, and X.~Li, ``{CNNs-based RGB-D saliency
  detection via cross-view transfer and multiview fusion},'' \emph{IEEE
  transactions on cybernetics}, vol.~48, no.~11, pp. 3171--3183, 2017.

\bibitem{mv-paper10}
\BIBentryALTinterwordspacing
G.~Kang, K.~Liu, B.~Hou, and N.~Zhang, ``{3D multi-view convolutional neural
  networks for lung nodule classificationq},'' \emph{PLOS ONE}, vol.~12,
  no.~11, pp. 1--21, 11 2017. [Online]. Available:
  \url{https://doi.org/10.1371/journal.pone.0188290}
\BIBentrySTDinterwordspacing

\bibitem{mv-paper11}
M.~Elhoseiny, T.~El-Gaaly, A.~Bakry, and A.~M. Elgammal, ``{A Comparative
  Analysis and Study of Multiview CNN Models for Joint Object Categorization
  and Pose Estimation},'' in \emph{ICML}, 2016.

\bibitem{mv-paper12}
F.~P.~S. {Luus}, B.~P. {Salmon}, F.~{van den Bergh}, and B.~T.~J. {Maharaj},
  ``{Multiview Deep Learning for Land-Use Classification},'' \emph{IEEE
  Geoscience and Remote Sensing Letters}, vol.~12, no.~12, pp. 2448--2452, Dec
  2015.

\bibitem{mv-paper13}
A.~Dai and M.~Niessner, ``{3DMV: Joint 3D-Multi-View Prediction for 3D Semantic
  Scene Segmentation},'' in \emph{{The European Conference on Computer Vision
  (ECCV)}}, September 2018.

\bibitem{mv-paper14}
E.~Kalogerakis, M.~Averkiou, S.~Maji, and S.~Chaudhuri, ``{3D shape
  segmentation with projective convolutional networks},'' in \emph{{Proceedings
  of the IEEE Conference on Computer Vision and Pattern Recognition}}, 2017,
  pp. 3779--3788.

\bibitem{changedetect}
S.~{Saha}, F.~{Bovolo}, and L.~{Bruzzone}, ``{Unsupervised Multiple-Change
  Detection in VHR Multisensor Images Via Deep-Learning Based Adaptation},'' in
  \emph{{IGARSS 2019 - 2019 IEEE International Geoscience and Remote Sensing
  Symposium}}, 2019, pp. 5033--5036.

\bibitem{CycleGan2017}
{Zhu, Jun-Yan and Park, Taesung and Isola, Phillip and Efros, Alexei A},
  ``{Unpaired Image-to-Image Translation using Cycle-Consistent Adversarial
  Networks},'' in \emph{Computer Vision (ICCV), 2017 IEEE International
  Conference on}, 2017.

\bibitem{dfc_winner}
P.~{d'Angelo}, D.~{Cerra}, S.~M. {Azimi}, N.~{Merkle}, J.~{Tian}, S.~{Auer},
  M.~{Pato}, R.~{de los Reyes}, X.~{Zhuo}, K.~{Bittner}, T.~{Krauss}, and
  P.~{Reinartz}, ``{3D Semantic Segmentation from Multi-View Optical Satellite
  Images},'' in \emph{IGARSS 2019 - 2019 IEEE International Geoscience and
  Remote Sensing Symposium}, July 2019, pp. 5053--5056.

\bibitem{dfc2019}
B.~Le~Saux, N.~Yokoya, R.~Hansch, M.~Brown, and G.~Hager, ``{2019 data fusion
  contest [technical committees]},'' \emph{IEEE Geoscience and Remote Sensing
  Magazine}, vol.~7, no.~1, pp. 103--105, 2019.

\bibitem{danesfield}
M.~J. Leotta, C.~Long, B.~Jacquet, M.~Zins, D.~Lipsa, J.~Shan, B.~Xu, Z.~Li,
  X.~Zhang, S.-F. Chang, M.~Purri, J.~Xue, and K.~Dana, ``{Urban Semantic 3D
  Reconstruction From Multiview Satellite Imagery},'' in \emph{The IEEE
  Conference on Computer Vision and Pattern Recognition (CVPR) Workshops}, June
  2019.

\bibitem{ohio_3d}
\BIBentryALTinterwordspacing
R.~Qin, ``{Automated 3D recovery from very high resolution multi-view satellite
  images},'' \emph{CoRR}, vol. abs/1905.07475, 2019. [Online]. Available:
  \url{http://arxiv.org/abs/1905.07475}
\BIBentrySTDinterwordspacing

\bibitem{largescale_isprs}
\BIBentryALTinterwordspacing
G.~Kuschk, ``{LARGE SCALE URBAN RECONSTRUCTION FROM REMOTE SENSING IMAGERY},''
  \emph{ISPRS - International Archives of the Photogrammetry, Remote Sensing
  and Spatial Information Sciences}, vol. XL-5/W1, pp. 139--146, 2013.
  [Online]. Available:
  \url{https://www.int-arch-photogramm-remote-sens-spatial-inf-sci.net/XL-5-W1/139/2013/}
\BIBentrySTDinterwordspacing

\bibitem{nasa-ames}
D.~Shean, O.~Alexandrov, Z.~Moratto, B.~Smith, I.~Joughin, C.~Porter, and
  P.~Morin, ``\BIBforeignlanguage{English (US)}{{An automated, open-source
  pipeline for mass production of digital elevation models (DEMs) from
  very-high-resolution commercial stereo satellite imagery}},''
  \emph{\BIBforeignlanguage{English (US)}{ISPRS Journal of Photogrammetry and
  Remote Sensing}}, vol. 116, pp. 101--117, 6 2016.

\bibitem{ozge_comparison}
O.~C. Ozcanli, Y.~Dong, J.~L. Mundy, H.~Webb, R.~Hammoud, and V.~Tom, ``{A
  comparison of stereo and multiview 3-D reconstruction using cross-sensor
  satellite imagery},'' in \emph{{2015 IEEE Conference on Computer Vision and
  Pattern Recognition Workshops (CVPRW)}}, June 2015, pp. 17--25.

\bibitem{s2p}
G.~Facciolo, C.~D. Franchis, and E.~Meinhardt-Llopis, ``{Automatic 3D
  Reconstruction from Multi-date Satellite Images},'' in \emph{{2017 IEEE
  Conference on Computer Vision and Pattern Recognition Workshops (CVPRW)}},
  July 2017, pp. 1542--1551.

\bibitem{demsculpt}
\BIBentryALTinterwordspacing
S.~Patil, T.~Prakash, B.~Comandur, and A.~C. Kak, ``{A Comparative Evaluation
  of {SGM} Variants (including a New Variant, tMGM) for Dense Stereo
  Matching},'' \emph{CoRR}, vol. abs/1911.09800, 2019. [Online]. Available:
  \url{http://arxiv.org/abs/1911.09800}
\BIBentrySTDinterwordspacing

\bibitem{isprs_ptcld_example}
\BIBentryALTinterwordspacing
K.~Gong and D.~Fritsch, ``{POINT CLOUD AND DIGITAL SURFACE MODEL GENERATION
  FROM HIGH RESOLUTION MULTIPLE VIEW STEREO SATELLITE IMAGERY},'' \emph{ISPRS -
  International Archives of the Photogrammetry, Remote Sensing and Spatial
  Information Sciences}, vol. XLII-2, pp. 363--370, 2018. [Online]. Available:
  \url{https://www.int-arch-photogramm-remote-sens-spatial-inf-sci.net/XLII-2/363/2018/}
\BIBentrySTDinterwordspacing

\bibitem{Pleiades}
\BIBentryALTinterwordspacing
R.~Perko, H.~Raggam, and P.~M. Roth, ``{Mapping with Pl\'eiades--End-to-End
  Workflow},'' \emph{Remote Sensing}, vol.~11, no.~17, 2019. [Online].
  Available: \url{https://www.mdpi.com/2072-4292/11/17/2052}
\BIBentrySTDinterwordspacing

\bibitem{rvlstereo}
\BIBentryALTinterwordspacing
S.~Patil, B.~Comandur, T.~Prakash, and A.~C. Kak, ``{A New Stereo Benchmarking
  Dataset for Satellite Images},'' \emph{CoRR}, vol. abs/1907.04404, 2019.
  [Online]. Available: \url{http://arxiv.org/abs/1907.04404}
\BIBentrySTDinterwordspacing

\bibitem{chipcluster}
R.~S. {Gargees} and G.~J. {Scott}, ``{Deep Feature Clustering for Remote
  Sensing Imagery Land Cover Analysis},'' \emph{IEEE Geoscience and Remote
  Sensing Letters}, vol.~17, no.~8, pp. 1386--1390, 2020.

\bibitem{rothermel2012sure}
M.~Rothermel, ``{Development of a {SGM}-based multi-view reconstruction
  framework for aerial imagery},'' Dissertation, University of Stuttgart, 2016.

\bibitem{unet}
\BIBentryALTinterwordspacing
O.~Ronneberger, P.Fischer, and T.~Brox, ``{U-Net: Convolutional Networks for
  Biomedical Image Segmentation},'' in \emph{Medical Image Computing and
  Computer-Assisted Intervention (MICCAI)}, ser. LNCS, vol. 9351.\hskip 1em
  plus 0.5em minus 0.4em\relax Springer, 2015, pp. 234--241, (available on
  arXiv:1505.04597 [cs.CV]). [Online]. Available:
  \url{http://lmb.informatik.uni-freiburg.de/Publications/2015/RFB15a}
\BIBentrySTDinterwordspacing

\bibitem{batchnorm}
\BIBentryALTinterwordspacing
S.~Ioffe and C.~Szegedy, ``{Batch Normalization: Accelerating Deep Network
  Training by Reducing Internal Covariate Shift},'' \emph{CoRR}, vol.
  abs/1502.03167, 2015. [Online]. Available:
  \url{http://arxiv.org/abs/1502.03167}
\BIBentrySTDinterwordspacing

\bibitem{deeplabv3}
Y.~Zhu, K.~Sapra, F.~A. Reda, K.~J. Shih, S.~Newsam, A.~Tao, and B.~Catanzaro,
  ``{Improving Semantic Segmentation via Video Propagation and Label
  Relaxation},'' in \emph{Proceedings of the IEEE Conference on Computer Vision
  and Pattern Recognition}, 2019, pp. 8856--8865.

\bibitem{Cordts2016Cityscapes}
M.~Cordts, M.~Omran, S.~Ramos, T.~Rehfeld, M.~Enzweiler, R.~Benenson,
  U.~Franke, S.~Roth, and B.~Schiele, ``{The Cityscapes Dataset for Semantic
  Urban Scene Understanding},'' in \emph{{Proc. of the IEEE Conference on
  Computer Vision and Pattern Recognition (CVPR)}}, 2016.

\bibitem{Grodecki}
\BIBentryALTinterwordspacing
J.~Grodecki and G.~Dial, ``{Block Adjustment of High-Resolution Satellite
  Images Described by Rational Polynomials},'' \emph{Photogrammetric
  Engineering \& Remote Sensing}, vol.~69, no.~1, pp. 59--68, 2003. [Online].
  Available:
  \url{https://www.ingentaconnect.com/content/asprs/pers/2003/00000069/00000001/art00004}
\BIBentrySTDinterwordspacing

\bibitem{oh2010piecewise}
J.~Oh, W.~H. Lee, C.~K. Toth, D.~A. Grejner-Brzezinska, and C.~Lee, ``{A
  piecewise approach to epipolar resampling of pushbroom satellite images based
  on RPC},'' \emph{Photogrammetric Engineering \& Remote Sensing}, vol.~76,
  no.~12, pp. 1353--1363, 2010.

\bibitem{zbontar2016stereo}
J.~Zbontar and Y.~LeCun, ``{Stereo matching by training a convolutional neural
  network to compare image patches},'' \emph{Journal of Machine Learning
  Research}, vol.~17, pp. 1--32, 2016.

\bibitem{srtm}
``{SRTM DEM},'' \url{https://www2.jpl.nasa.gov/srtm/}, accessed: 2018-08-06.

\bibitem{nhfd}
\BIBentryALTinterwordspacing
A.~F. Wolf, ``{Using WorldView-2 Vis-NIR multispectral imagery to support land
  mapping and feature extraction using normalized difference index ratios},''
  in \emph{Algorithms and Technologies for Multispectral, Hyperspectral, and
  Ultraspectral Imagery XVIII}, S.~S. Shen and P.~E. Lewis, Eds., vol. 8390,
  International Society for Optics and Photonics.\hskip 1em plus 0.5em minus
  0.4em\relax SPIE, 2012, pp. 188 -- 195. [Online]. Available:
  \url{https://doi.org/10.1117/12.917717}
\BIBentrySTDinterwordspacing

\bibitem{nhfd_2}
\BIBentryALTinterwordspacing
T.~Prakash and A.~C. Kak, ``{Active learning for designing detectors for
  infrequently occurring objects in wide-area satellite imagery},''
  \emph{Computer Vision and Image Understanding}, vol. 170, pp. 92 -- 108,
  2018. [Online]. Available:
  \url{http://www.sciencedirect.com/science/article/pii/S1077314218300390}
\BIBentrySTDinterwordspacing

\bibitem{csf}
\BIBentryALTinterwordspacing
W.~Zhang, J.~Qi, P.~Wan, H.~Wang, D.~Xie, X.~Wang, and G.~Yan, ``{An
  Easy-to-Use Airborne LiDAR Data Filtering Method Based on Cloth
  Simulation},'' \emph{Remote Sensing}, vol.~8, no.~6, p. 501, Jun 2016.
  [Online]. Available: \url{http://dx.doi.org/10.3390/rs8060501}
\BIBentrySTDinterwordspacing

\bibitem{openmp}
``{OpenMP},'' \url{https://www.openmp.org/}, accessed: 2020-05-20.

\end{thebibliography}
%



%

\begin{IEEEbiography}[{\includegraphics[width=1in,height=1.25in,clip,keepaspectratio]{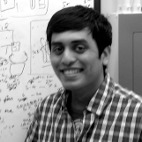}}]{Bharath
    Comandur} received the Bachelor of Technology degree in
  Electrical Engineering from the Indian Institute of
  Technology, Madras in 2012 and obtained his Ph.D. in
  Computer Engineering from the school of Electrical and
  Computer Engineering, Purdue University where he was
  affiliated with the Robot Vision Laboratory. His research
  interests include computer vision, machine learning,
  remote sensing and cloud computing.
\end{IEEEbiography}

\vspace{-2cm}
\begin{IEEEbiography}[{\includegraphics[width=1in,height=1.25in,clip,keepaspectratio]{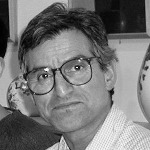}}]{Avinash C.
    Kak}
  is a professor of electrical and computer engineering at
  Purdue University. His research focuses on algorithms,
  languages, and systems related to wired and wireless
  camera networks, robotics, computer vision, etc. His
  coauthored book Principles of Computerized Tomographic
  Imaging was republished as a classic in applied
  mathematics by the Society of Industrial and Applied
  Mathematics. His other coauthored book, Digital Picture
  Processing, also considered by many to be a classic in
  computer vision and image processing, was published by
  Academic Press in 1982. His more recent books were written
  for his ``Objects Trilogy'' project. All three books of
  the Trilogy have been published by John Wiley \& Sons. the
  first, Programming with Objects, came out in 2003, the
  second, Scripting with Objects, in 2008, and the last,
  Designing with Objects, in 2015.
\end{IEEEbiography}








\end{document}